\documentclass[10pt,twocolumn,letterpaper]{article}

\PassOptionsToPackage{table}{xcolor} 
\usepackage[pagenumbers]{cvpr} 
\usepackage{multirow}
\usepackage{adjustbox}
\usepackage{marvosym} 
\usepackage{booktabs}       
\usepackage[table]{xcolor}  

\providecommand{\tablebody}{%
  \normalfont\footnotesize
  \setlength{\tabcolsep}{2pt}%
  \renewcommand{\arraystretch}{1.12}%
}

\definecolor{cvprblue}{rgb}{0.21,0.49,0.74}
\usepackage[pagebackref,breaklinks,colorlinks,allcolors=cvprblue]{hyperref}

\makeatletter
\let\rawmoment@fnsymbol\@fnsymbol
\renewcommand{\@fnsymbol}[1]{%
  \ifnum#1=1\relax\Letter\else\rawmoment@fnsymbol{#1}\fi
}
\makeatother

\def\confName{CVPR}
\def\confYear{2027}

\title{RawMoment: Recovering the Sharp Moment \\ via Generative RAW-to-RGB Deblurring}

\author{%
Sicheng Yang\textsuperscript{1}\quad
Song Fei\textsuperscript{1}\quad
Tian Ye\textsuperscript{1}\quad
Jianan Sun\textsuperscript{2}\\
Bin Chen\textsuperscript{2}\quad
Yu Wang\textsuperscript{2}\quad
Lei Zhu\textsuperscript{1,3}\thanks{Corresponding author.}\\[0.5em]
\textsuperscript{1}The Hong Kong University of Science and Technology (Guangzhou)\\
\textsuperscript{2}Honor Device Co., Ltd.\\
\textsuperscript{3}The Hong Kong University of Science and Technology\\[0.3em]
{\tt\small syang671@connect.hkust-gz.edu.cn\quad leizhu@hkust-gz.edu.cn}
}

\begin{document}
\maketitle
\begin{abstract}
Jointly performing motion deblurring and image signal processing on RAW measurements is challenging. Progress is limited by insufficient scene, motion, and camera diversity in paired datasets, while existing methods have yet to fully exploit pretrained generative priors for this joint task. We present \textbf{RawMoment}, which brings together motion supervision from real videos and the visual knowledge of a pretrained text-to-image model. Our data construction separates natural motion collection from sensor simulation. Carefully screened video segments provide reliable sharp references and varied motion, while temporal integration and calibrated RAW rendering produce paired supervision under controllable acquisition conditions. This process also establishes RawMoment-Bench for evaluation on unseen source videos. To make the pretrained image prior usable for RAW restoration, we learn a sensor representation compatible with its RGB latent space and guide generation with spatial observations and physical acquisition parameters. Trained entirely on synthetic pairs, RawMoment generalizes to external real captures across cameras without camera-specific fine-tuning. Across RawMoment-Bench, RealBlur, and Magic8-Pro, it achieves superior perceptual quality, with an average relative MUSIQ improvement of 33.2\% over the strongest competing method on each dataset. Code and dataset: \url{https://github.com/script-Yang/rawmoment}.
\end{abstract}    
\section{Introduction}
\label{sec:intro}

\begin{figure}[!t]
    \centering
    \includegraphics[width=\linewidth]{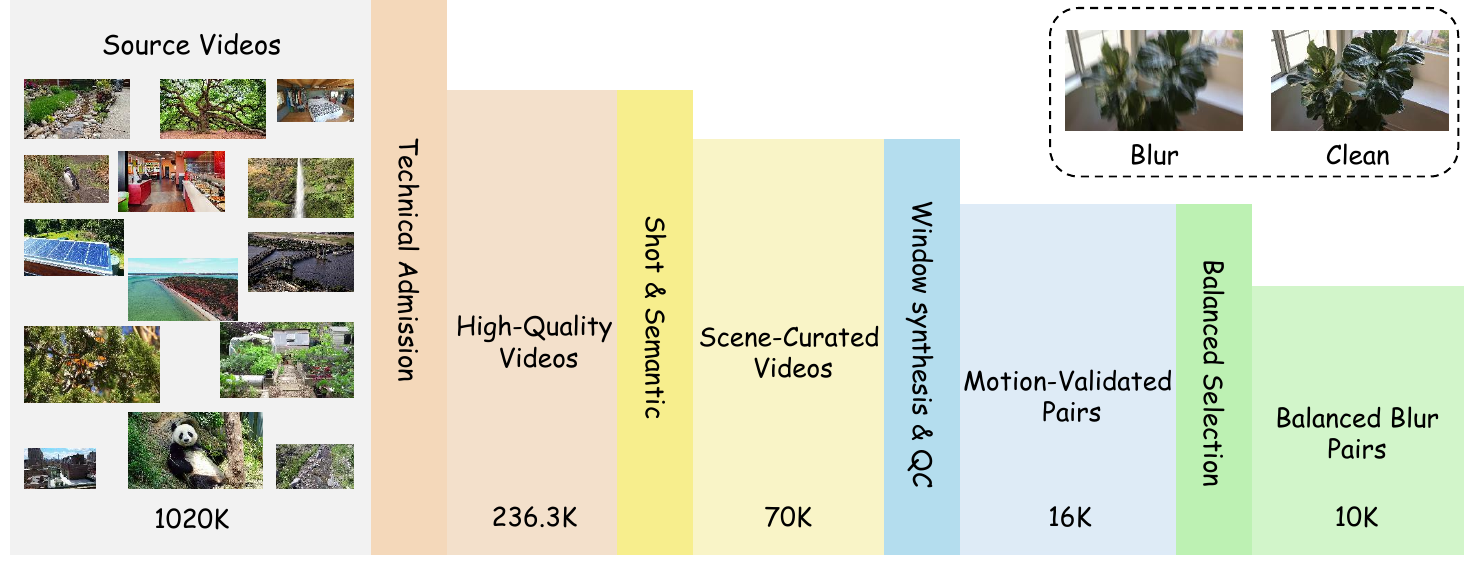}
    \caption{\textbf{RawMoment data construction pipeline.}
    We curate OpenVid-1M videos through a series of rule-based and
    semantic filters, then integrate frames in linear light to form
    sharp/blurred pairs.
    Window-level quality control, deduplication, and blur-strength
    balancing yield 10,000 RGB pairs for RAW synthesis.}
    \label{fig:data_pipeline}
\end{figure}

Motion blur is a frequent problem in
photography~\citep{nah2017deep,rim2020realblur}.
In low light, cameras often extend exposure to collect enough photons,
allowing camera shake and object motion to accumulate.
Sensor noise further obscures edges and textures, degrading visual
quality~\citep{chen2018learning,rim2020realblur}.
Recovering a sharp, well-rendered image therefore requires
deblurring, denoising, and color reconstruction to work together.
However, jointly performing deblurring and RAW image signal
processing (ISP) remains challenging~\citep{liang2020raw,jin2023dnf}.

Benchmarks such as GoPro~\citep{nah2017deep} and
RealBlur~\citep{rim2020realblur} have advanced learned deblurring. Recently, approaches such as RAWNet~\citep{liang2020raw} and DNF~\citep{jin2023dnf} explore deblurring
and low-light RAW-to-RGB processing.
Nevertheless, two limitations remain.
First, existing datasets provide limited joint coverage of scene
semantics, motion, and sensor conditions.
GoPro contains 3,214 RGB pairs from a single camera model and
offers no paired RAW--RGB supervision for joint ISP and
deblurring~\citep{nah2017deep}.
Although RealBlur provides RAW images, its scene diversity
remains limited, and the high cost of real-world data
acquisition makes large-scale expansion
challenging~\citep{rim2020realblur}.
Second, large pretrained text-to-image (T2I) priors remain
underused for joint RAW ISP and deblurring.
Pretrained T2I models offer image statistics learned from much broader
data, with demonstrated perceptual benefits in RGB
restoration~\citep{yu2024scaling,sun2025pixel}.
LDM-ISP~\citep{wen2025ldm} already transfers such a prior to low-light
RAW enhancement, but focuses on exposure recovery.
Extending this capability to joint RAW ISP and deblurring requires a
bridge between RAW observations and RGB generation that preserves
motion-degraded spatial and sensor information.

In this paper, we propose \textbf{RawMoment}, which couples
high-quality data construction with a pretrained T2I model for joint
RAW ISP and deblurring.
As shown in \cref{fig:data_pipeline}, our data construction pipeline selects reliable
motion segments and sharp references from diverse real videos,
balancing scene semantics and blur strength.
We combine linear-light temporal integration with calibrated sensor
synthesis to produce RAW--RGB supervision under varied acquisition
conditions.
To exploit pretrained T2I priors, we first adapt a RAW encoder to the
model's RGB latent space, addressing the representation mismatch
between sensor measurements and RGB images.
We then condition the pretrained T2I model on spatial features from
the RAW input and sensor parameters, guiding sharp RGB reconstruction
while preserving spatial correspondence and accounting for sensor
conditions.
Trained for restoration using only our synthetic pairs, RawMoment
achieves state-of-the-art perceptual quality against representative
baselines and generalizes to real captures across cameras and blur
distributions without camera-specific fine-tuning.

Our contributions are threefold:
\begin{itemize}
    \item A RAW data construction pipeline that integrates
    video curation, quality filtering, and calibrated sensor
    rendering to generate motion-blurred RAW inputs paired
    with sharp RGB references for joint ISP and deblurring.
    We also establish RawMoment-Bench for held-out evaluation.

    \item A unified framework for joint RAW ISP and deblurring
    that adapts pretrained T2I priors to sensor RAW observations
    through RAW-to-RGB latent alignment and spatial and
    physical conditioning.

    \item Extensive experiments demonstrate that our method
    reduces LPIPS by 37.6\% on RawMoment-Bench and 18.3\%
    on RealBlur compared with the best competing methods,
    while generalizing across cameras without camera-specific
    fine-tuning.
    A pilot user study further achieves the highest overall
    mean opinion score of 4.62, exceeding the
    best competing method by 1.22 points.
\end{itemize}
\section{Related Work}
\label{sec:related}

\paragraph{RAW ISP and Deblurring.}
Learned RAW pipelines couple low-light recovery with RGB reconstruction,
as in SID~\citep{chen2018learning} and DNF~\citep{jin2023dnf},
or adapt ISP to downstream perception through
RAW-Adapter~\citep{cui2024rawadapter}.
Within photographic restoration, motion-degraded captures additionally
require deblurring, explored in the RAW
domain~\citep{liang2020raw,jiao2025efficient} and jointly with
denoising~\citep{wang2026extreme}.
Approaches such as RSBlur~\citep{rim2022realistic} improve synthetic
supervision by modeling noise and ISP effects, reducing the gap
to real blurred images.
However, degradation realism alone does not guarantee diverse
scene and motion coverage or reliable sharp references.
Our work combines quality-controlled real-video motion and sharp
references with calibrated sensor synthesis, and uses rich
pretrained T2I priors for joint RAW ISP and deblurring.

\paragraph{Pretrained Generative Priors for Image Restoration.}
Pretrained T2I models improve perceptual quality in RGB restoration
through conditional adaptation~\citep{yu2024scaling,deng2025acquire,fei2026lucidflux}.
In RAW imaging, task-specific generative ISP learns texture and
color mappings~\citep{ren2025ispdiffuser}; transferring a pretrained
T2I prior further requires reconciling sensor and RGB representations.
LDM-ISP~\citep{wen2025ldm} bridges this gap through feature modulation
of frozen Stable Diffusion for low-light enhancement, whereas
RDDM~\citep{chen2025rddm} adapts a VAE for RAW-to-linear restoration
with Bayer-specific LoRA.
We focus on joint RAW ISP and deblurring by aligning observations
to the pretrained RGB latent space, retaining the RGB decoder, and
conditioning a pretrained diffusion transformer on spatial evidence
and sensor parameters.

\section{Method}
\label{sec:method}

\begin{figure*}[t]
    \centering
    \includegraphics[width=\textwidth]{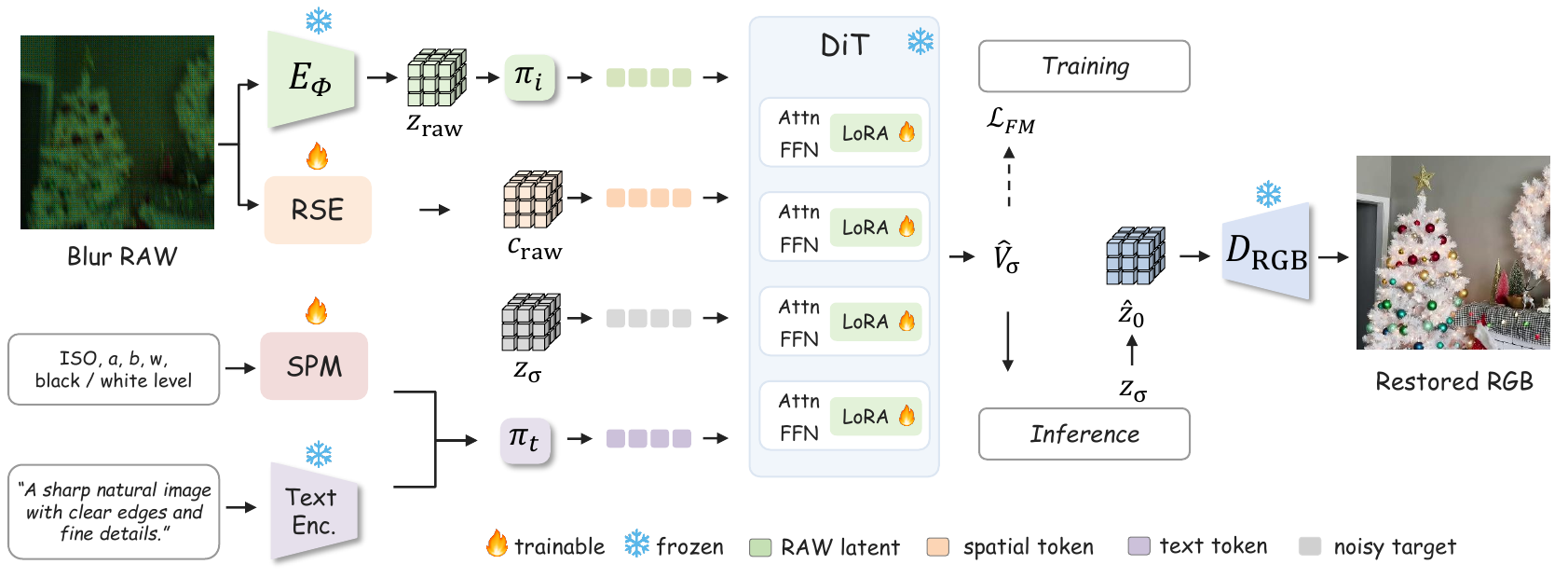}
    \caption{\textbf{Overview of RawMoment for joint RAW ISP and deblurring.}
    The adapted encoder $E_{\Phi}$ maps the blurred RAW input to the
    pretrained RGB latent space.
    RAW Spatial Encoding (RSE) extracts spatial structure from RAW, while
    Sensor Physics Modulation (SPM) encodes physical sensor parameters.
    These conditions guide restoration through the pretrained T2I model's
    DiT blocks, which are adapted with attention LoRA.
    RSE, SPM, and the LoRA adapters are jointly trained with flow matching
    to predict the velocity field.
    At inference, the model starts from Gaussian noise and performs
    iterative Euler updates conditioned on the blurred RAW input to
    obtain $\hat z_0$, which the frozen RGB decoder $D_{\mathrm{RGB}}$
    converts to the restored image.}
    \label{fig:model}
\end{figure*}

We first construct paired RAW--RGB supervision from curated real
videos through linear-light temporal integration and calibrated sensor
synthesis.
Using these pairs, we train RawMoment in two stages, as shown in
\cref{fig:model,fig:latent_alignment}.
The first stage adapts a RAW encoder to the pretrained RGB latent
space.
In the second stage, RAW Spatial Encoding (RSE) and Sensor Physics
Modulation (SPM) complement the adapted latent with spatial features
and sensor conditions, respectively.
These conditions jointly guide the pretrained T2I model, adapted
through LoRA, to reconstruct a sharp RGB image from a
blurred RAW input.

\subsection{RawMoment Data Construction}
\label{sec:data}
Paired RAW data with real motion blur remain scarce. Existing RAW deblurring datasets are typically captured with a single camera and cover a limited range of scenes~\citep{liang2020raw,rim2020realblur}. In this paper, we develop a scalable data synthesis pipeline to expand the scale and diversity of paired RAW training data. Real-world videos capture natural camera and object motion across diverse scenes. By temporally integrating consecutive frames, we simulate motion blur caused by long exposure while using the center frame as the sharp reference. This allows us to construct diverse sharp/blurred pairs without requiring paired RAW captures. We further demonstrate the generalization of models trained on these synthetic data to real-world images captured by different cameras. As shown in \cref{fig:data_pipeline}, we first construct sharp-blurred RGB pairs from real videos and then render them into sensor RAW using a calibrated degradation model.

\paragraph{Real-video Curation.}
Our source pool is OpenVid-1M~\citep{nan2025openvid}, which contains
1,019,957 short real-world video clips.
Real videos cover far more scenes than multi-frame capture rigs such as
GoPro~\citep{nah2017deep}, but they come without paired ground truth.
We synthesize blur from consecutive frames and filter the pool in
four stages.
(i)~\textit{Technical admission} keeps clips that decode at
$\geq$1920$\times$1080 with SDR BT.709 color~\citep{itu2015bt709}, leaving 236,276 clips.
(ii)~\textit{Shot and semantic admission} discards clips with any
detected cut. We then use Qwen3-VL-8B-Instruct~\citep{bai2025qwen3vl}
to inspect five frames sampled across each clip; the model rejects
non-camera content (CG, screen recordings, and interfaces) and assigns
one of ten content classes.
Synthesis candidates are then drawn from these classes under equal
per-class quotas, so that frequent content such as talking people does
not dominate the set.
(iii)~\textit{Window synthesis and quality control.} We estimate
forward and backward optical flow between adjacent frames with
SEA-RAFT~\citep{wang2024searaft} at a long side of 960 pixels.
A blurred frame is formed from $N{=}15$ consecutive frames
$\{I_t\}$ by averaging in linear light,
\begin{equation}
y_b = f\!\left(\frac{1}{N}\sum_{t=1}^{N} f^{-1}(I_t)\right),
\qquad y_c = I_{\lceil N/2\rceil},
\label{eq:temporal_average}
\end{equation}
where $f$ is the BT.709 transfer function and the center frame $y_c$
serves as the sharp target.
A window is rejected if it contains a cut, an exposure jump,
irregular timestamps (more than 20\% deviation from the median frame
interval), or forward--backward flow inconsistency.
It is also rejected if the 95th-percentile adjacent displacement
exceeds 8 pixels, a proxy for temporal undersampling that would
produce ghosting instead of continuous blur, or if the center frame
fails sharpness and exposure checks.
We further require the blurred frame to be visibly less sharp than
$y_c$, and measure blur strength as the accumulated flow magnitude
normalized to a 1080-pixel reference height.
Windows with strength in $[5,80)$ are kept and grouped into four
buckets.
Of 3.35M candidate windows, 16,265 pass all checks.
(iv)~\textit{Selection} keeps at most three temporally non-overlapping
windows per original video, removes exact duplicates, and balances the
strength buckets.

The dataset comprises 10,000 pairs derived from 9,142 distinct videos, 
with 97.9\% presented in 1920$\times$1080 resolution and the remainder in 4K or 8K. 
Exposure windows range from 233--584\,ms. 
The four levels contain 1,652, 3,430, 3,384, and 1,534 pairs
in order of increasing blur strength.
Local blur, caused by moving objects or regions, accounts for 65.5\% of the pairs, 
while global blur due to camera motion or frame-filling subjects comprises 34.5\%.

\paragraph{Sensor RAW Synthesis.}
For each pair, we take a 1024$\times$1024 center crop and render the blurred frame into a low-light RAW observation. Starting from $y_b$, we reduce brightness and apply inverse white-balance gains while reserving headroom for the black level $\beta$:
\begin{equation}
u = \mathbf{w}\odot (1-\beta)\,y_b^{\gamma},
\qquad
\mathbf{w}=(g_R^{-1},1,g_B^{-1}),
\label{eq:raw_forward}
\end{equation}
where $\gamma$ controls the brightness attenuation. Since the noise coefficients $(a,b)$ are calibrated on RAW measurements before white-balance amplification, we inject heteroscedastic sensor noise in this domain:
\begin{equation}
\tilde u = \operatorname{clip}\!\left(u+n,\,0,\,1-\beta\right),
\qquad n\sim\mathcal{N}\!\left(0,\;a\,u+b\right).
\label{eq:raw_noise}
\end{equation}
The synthesis parameters and ISO-dependent noise calibration are detailed in the supplementary material. After noise injection, we restore the white-balance gains to map the noisy signal back to its original scale, restore the black level, and sample the result on an RGGB Bayer pattern:
\begin{equation}
r_b = \mathcal{M}_{\mathrm{RGGB}}\!\left(
\operatorname{clip}(\tilde u\oslash\mathbf{w},\,0,\,1-\beta)+\beta
\right),
\label{eq:raw_mosaic}
\end{equation}
which is stored at 16 bits. The sharp frame $y_c$ passes through the same operator with the same ISO, gains, and noise seed to give a clean RAW $r_c$. Here clean means free of motion blur; $r_c$ still contains the low-light degradation and sensor noise.

\paragraph{RawMoment Dataset.}
We construct a dataset comprising 10,000 sharp/blurred
pairs curated from 9,142 real-world videos.
Each pair is rendered into three RAW variants with
independently sampled sensor parameters, yielding
30,000 paired RAW-RGB tuples.
The dataset is split by source video into 29,634 training
and 366 validation samples, ensuring no video overlap.
For evaluation, we additionally construct \textbf{RawMoment-Bench}
with 200 pairs from held-out OpenVid videos, disjoint
from the training and validation sets.
The benchmark covers 12 balanced scene classes and follows the same RAW synthesis pipeline.

\subsection{RAW-to-RGB Latent Adaptation}
\label{sec:latent_alignment}
\begin{figure}[t]
    \centering
    \includegraphics[width=\linewidth]{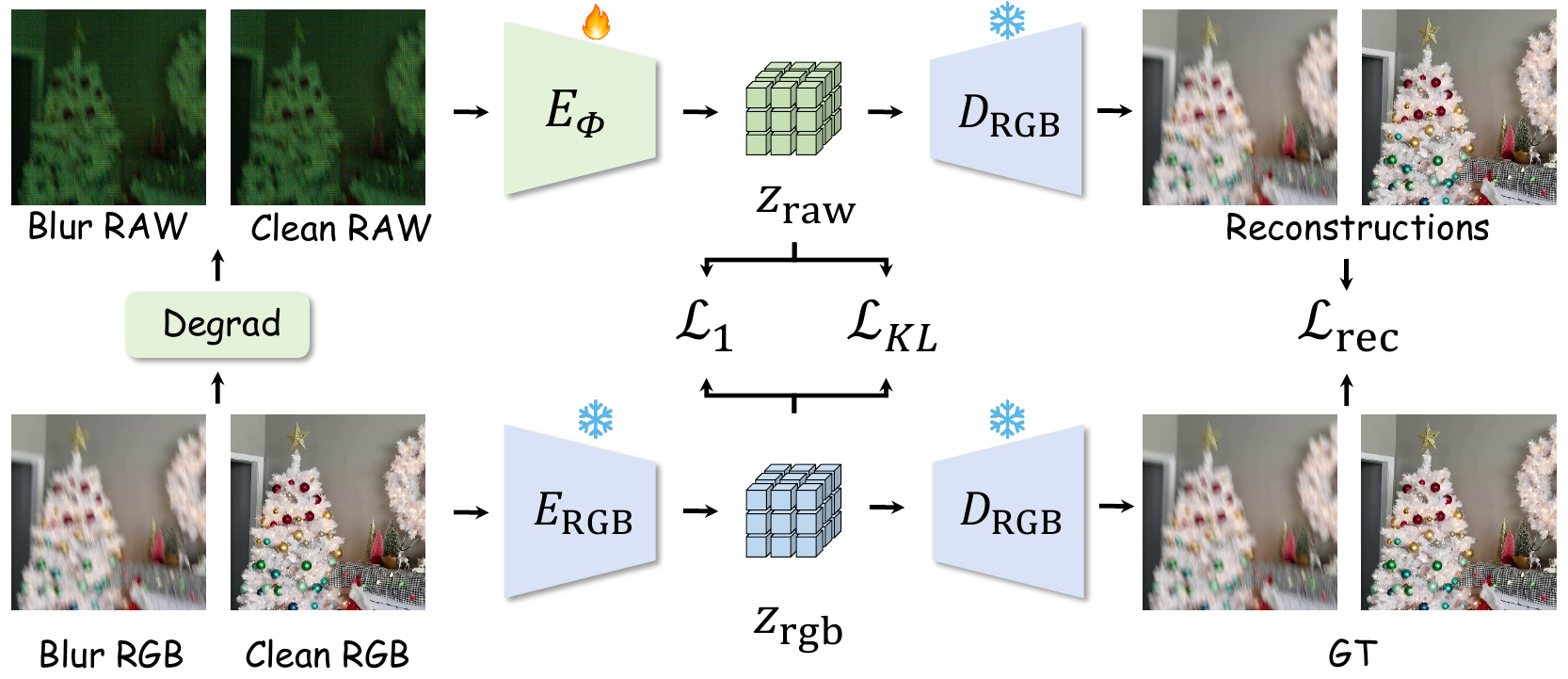}
    \caption{\textbf{RAW-to-RGB latent adaptation.}
    A shared RAW encoder $E_{\Phi}$ is adapted on clean and blurred inputs.
    Latent losses $\mathcal{L}_1$ and $\mathcal{L}_{\mathrm{KL}}$
    align its outputs with the frozen RGB encoder, while
    $\mathcal{L}_{\mathrm{rec}}$ supervises reconstructions through
    the frozen RGB decoder.}
    \label{fig:latent_alignment}
\end{figure}

We adapt a RAW encoder $E_{\Phi}$ to the pretrained RGB VAE latent
space so that sensor observations can condition RGB generation.
As shown in \cref{fig:latent_alignment}, adaptation uses both clean
and blurred pairs $(r_d,y_d)$, $d\in\{c,b\}$, from \cref{sec:data}.
The shared encoder is initialized from $E_{\mathrm{RGB}}$ and
accepts sparse four-channel R/G1/G2/B inputs.
The RGB encoder $E_{\mathrm{RGB}}$ and decoder $D_{\mathrm{RGB}}$
remain frozen.

For each pair, the encoders yield
$z_{\mathrm{raw}}^{d}=E_{\Phi}(r_d)$ and
$z_{\mathrm{rgb}}^{d}=E_{\mathrm{RGB}}(y_d)$ using the same VAE
latent normalization.
We match the latent means with $\mathcal{L}_1^d$ and use
$\mathcal{L}_{\mathrm{KL}}^d$ to align the RAW encoding statistics
with the RGB reference.
Decoding the sampled RAW latent gives
$\hat y_d=D_{\mathrm{RGB}}(z_{\mathrm{raw}}^{d})$, supervised by
\begin{equation}
\begin{aligned}
\mathcal{L}_{\mathrm{rec}}^d
&=\|\hat y_d-y_d\|_1\\
&\quad+\lambda_{\mathrm{p}}\,
\operatorname{LPIPS}_{\mathrm{VGG}}(\hat y_d,y_d),
\end{aligned}
\label{eq:domain_alignment}
\end{equation}
where $\lambda_{\mathrm{p}}=0.1$ and LPIPS uses a frozen VGG
backbone~\cite{zhang2018unreasonable}.
The adaptation objective averages the losses over both blur states:
\begin{equation}
\mathcal{L}_{\mathrm{adapt}}
=\frac{1}{2}\sum_{d\in\{c,b\}}
\left(\mathcal{L}_1^d+\mathcal{L}_{\mathrm{KL}}^d
+\mathcal{L}_{\mathrm{rec}}^d\right).
\label{eq:alignment_objective}
\end{equation}
Only $E_{\Phi}$ is optimized, with reconstruction gradients
propagating through the frozen decoder.
After adaptation, we freeze $E_{\Phi}$ and use its latent mean
$z_{\mathrm{raw}}$ as the spatial condition for restoration.
This adaptation bridges the representation gap between sensor
RAW and the latent space of the pretrained RGB generative model.

\subsection{Spatial and Physical Conditioning}
\label{sec:conditioning}

We complement the adapted latent with two trainable conditioning
branches for joint ISP and deblurring.
RSE extracts spatial features directly from the degraded RAW
input $r=r_b$, while SPM encodes the sensor acquisition settings.

\paragraph{RAW Spatial Encoding (RSE).}
Latent adaptation connects RAW observations to the pretrained RGB
space.
To supply spatial features that can be optimized for joint ISP and
deblurring, we introduce RSE as a trainable branch operating directly
on the RAW input.
The RSE encoder $G_{\psi}$ packs the normalized RGGB input by
Bayer phase and applies a residual CNN, producing a 64-channel
feature grid at $H/16\times W/16$ for an $H\times W$ input.
Each $2\times2$ neighborhood is concatenated into a patch vector.
A projection $\Pi_{\mathrm{s}}$, comprising LayerNorm followed by
a linear layer, maps each vector to the DiT hidden width:
\begin{equation}
C_{\mathrm{raw}}
=\Pi_{\mathrm{s}}\!\left(
\operatorname{Patch}_{2\times2}(G_{\psi}(r))
\right)+e_{\mathrm{raw}}.
\label{eq:raw_spatial_tokens}
\end{equation}
The learned embedding $e_{\mathrm{raw}}$ identifies the RAW stream.
The resulting tokens retain an $H/32\times W/32$ spatial grid,
preserving the correspondence between RAW regions and their
conditioning features.
This provides the DiT with a spatial reference from the sensor
observation, complementing the RGB-aligned latent when reconstructing
sharp edges and textures.

\paragraph{Sensor Physics Modulation (SPM).}
Sensor acquisition settings affect both the noise statistics of RAW
measurements and their conversion to RGB.
We introduce SPM to expose these conditions to the restoration model,
complementing the spatial information supplied by RSE.
We form a normalized parameter vector $\mathbf{p}$ from ISO,
noise coefficients $(a,b)$, white balance $\mathbf{w}$, and
black/white levels.
An MLP $P_{\omega}$ maps this vector to physical tokens:
\begin{equation}
C_{\mathrm{p}}=P_{\omega}(\mathbf{p}).
\label{eq:sensor_physics_tokens}
\end{equation}
Normalization and architecture details are given in the
supplementary material.
The resulting tokens provide a global description of the acquisition
conditions alongside the spatially organized RAW features.
This allows the DiT to interpret the spatial RAW evidence in the
context of the corresponding noise and color settings, guiding RGB
reconstruction under varied acquisition conditions.

\subsection{Conditional Generative Restoration}
\label{sec:generative_restoration}

As shown in \cref{fig:model}, the adapted RAW latent, RSE features,
and SPM tokens condition a pretrained DiT to generate a sharp RGB
latent from noise.

\paragraph{Condition Injection.}
We encode the fixed prompt ``A sharp natural image with clear edges
and fine details.'' once and cache the resulting text tokens
$C_{\mathrm{text}}$.
With the pretrained text and image projections
$\Pi_{\mathrm{t}}$ and $\Pi_{\mathrm{i}}$, we form the condition prefix:
\begin{equation}
\mathcal{C}=
\Big[
\Pi_{\mathrm{t}}([C_{\mathrm{text}};C_{\mathrm{p}}]);\,
C_{\mathrm{raw}};\,
\Pi_{\mathrm{i}}(z_{\mathrm{raw}})
\Big].
\label{eq:condition_sequence}
\end{equation}
To restore the noisy latent $z_{\sigma}$, the DiT predicts its
velocity from $[\mathcal{C};\Pi_{\mathrm{i}}(z_{\sigma})]$,
combining the adapted RAW latent with spatial features and
sensor information.
Spatially aligned rotary embeddings preserve correspondence
with the RAW observation, while block causal attention gives
the target access to the full condition prefix and keeps
condition features independent of the evolving target.

\paragraph{Training Objective.}
We train with flow matching~\cite{lipman2023flow}, using the normalized
latent mean $z_{\mathrm{rgb}}^c$ of the sharp RGB target $y_c$
from \cref{sec:latent_alignment}.
For $\epsilon\sim\mathcal{N}(0,I)$ and $\sigma\in[0,1]$ sampled
from the pretrained noise schedule, we form
\begin{equation}
z_{\sigma}
=
(1-\sigma)z_{\mathrm{rgb}}^c+\sigma\epsilon.
\label{eq:flow_interpolation}
\end{equation}
The adapted DiT predicts
$\hat v_{\sigma}=v_{\theta}(z_{\sigma},\sigma;\mathcal{C})$,
supervised by the interpolation velocity $\epsilon-z_{\mathrm{rgb}}^c$:
\begin{equation}
\mathcal{L}_{\mathrm{FM}}
=\mathbb{E}\!\left[
\operatorname{MSE}\!\left(
\hat v_{\sigma},\,
\epsilon-z_{\mathrm{rgb}}^c
\right)\right],
\label{eq:flow_matching}
\end{equation}
We apply LoRA~\cite{hu2022lora} to the DiT attention layers.
These adapters, RSE, and SPM are jointly optimized under
$\mathcal{L}_{\mathrm{FM}}$; all other parameters remain frozen.

\paragraph{Inference.}
We initialize $z_1\sim\mathcal{N}(0,I)$ and keep $\mathcal{C}$ fixed.
For a decreasing schedule
$1=\sigma_0>\sigma_1>\cdots>\sigma_K=0$,
we take $K$ Euler steps:
\begin{equation}
z_{\sigma_{k+1}}
=z_{\sigma_k}+\Delta\sigma_k\,\hat v_{\sigma_k},
\label{eq:euler_restoration}
\end{equation}
where $\Delta\sigma_k=\sigma_{k+1}-\sigma_k$.
The frozen RGB decoder maps the final latent $z_0$ to the restored image:
\begin{equation}
\hat y_c=D_{\mathrm{RGB}}(z_0).
\label{eq:restored_rgb}
\end{equation}

\subsection{Experimental Settings}
\label{sec:experimental_settings}

\subsubsection{Datasets}
\label{sec:datasets}
We train both stages on RawMoment pairs and evaluate the same trained model on RawMoment-Bench.
To assess whether the model trained on pairs synthesized from
our curated high-quality videos generalizes across cameras,
we further evaluate on two external out-of-distribution datasets, RealBlur and Magic8-Pro.
These datasets contain real RAW captures with different
camera responses and motion-blur characteristics.

\paragraph{RawMoment-Bench.}
RawMoment-Bench contains 200 RAW--RGB test pairs from videos
disjoint from training and validation, generated with the same
sensor synthesis pipeline.
We use fixed $512\times512$ center crops and physical conditions
from the synthesis records.

\paragraph{RealBlur.}
We assess cross-camera transfer on the 980 official
RealBlur-R~\citep{rim2020realblur} test pairs captured with a Sony A7R3.
We apply CFA-preserving binning to the captured Bayer RAW inputs
and use registered sharp RGB references.
Camera metadata and an ISO-dependent noise model supply the
physical conditions.
Evaluation uses fixed $512\times512$ crops within fully valid
paired regions.
Following the intensity-matching step of RealBlur
evaluation~\citep{rim2020realblur}, we align each prediction to
its reference with a single least-squares RGB brightness
scale before scoring.

\paragraph{Magic8-Pro.}
We collect 107 unpaired real RAW images using the main camera
of an HONOR Magic8 Pro at a fixed 23\,mm-equivalent focal length.
Captures span scene illuminance of 300--2000\,lux and exposure
times of 1--30\,ms.
For dynamic scenes, subjects move at 0.5--2\,m/s at
camera-to-subject distances of 0.5--3\,m.
We evaluate fixed $512\times512$ center crops, using recorded
ISO and white balance with ISO-dependent noise coefficients
for physical conditioning.

\begin{table*}[t]
\centering
\caption{Quantitative comparison on RawMoment-Bench,
RealBlur, and Magic8-Pro datasets.
The best and second-best results for each metric are highlighted
in \textbf{bold} and \underline{underlined}, respectively.
The same convention applies to all subsequent tables.}
\label{tab:blur_iqa}
\setlength{\tabcolsep}{3.3pt}
\renewcommand{\arraystretch}{1.05}

\resizebox{\textwidth}{!}{%
\begin{tabular}{l|l|cccccc|cccc}
\toprule

\multirow{2}{*}{\textbf{Benchmark}}
& \multirow{2}{*}{\textbf{Metric}}
& \multicolumn{6}{c|}{
    \cellcolor{green!12}\textbf{Traditional Methods}}
& \multicolumn{4}{c}{
    \cellcolor{blue!12}\textbf{Generative Methods}} \\

\cmidrule(lr){3-8}
\cmidrule(lr){9-12}

&
& \cellcolor{green!6}\textbf{SID}
& \cellcolor{green!6}\textbf{RAWNet}
& \cellcolor{green!6}\textbf{DeepRFT}
& \cellcolor{green!6}\textbf{DNF}
& \cellcolor{green!6}\textbf{LoFormer}
& \cellcolor{green!6}\textbf{RAWMamba}
& \cellcolor{blue!6}\textbf{LDM-ISP}
& \cellcolor{blue!6}\textbf{ISPDiffuser}
& \cellcolor{blue!6}\textbf{MeanFlow-RAW}
& \cellcolor{blue!6}\textbf{Ours} \\

\midrule

\multirow{10}{*}{\textbf{RawMoment-Bench}}
& CLIP-IQA+$\uparrow$
& 0.3565 & 0.3872 & 0.3807 & 0.3619
& 0.3888 & 0.3255 & \underline{0.4488}
& 0.3781 & 0.3106 & \textbf{0.4788} \\

& Q-Align$\uparrow$
& 2.3552 & 2.5257 & 2.6302 & 2.3360
& 2.6709 & 2.1585 & \underline{2.7727}
& 2.6011 & 2.0341 & \textbf{3.4549} \\

& MUSIQ$\uparrow$
& 34.9639 & 38.3033 & 39.8505 & 34.0458
& 41.7716 & 31.8195 & \underline{43.3717}
& 38.5115 & 34.1009 & \textbf{56.0882} \\

& MANIQA$\uparrow$
& 0.1736 & 0.2030 & 0.2090 & 0.1765
& \underline{0.2181} & 0.1627 & 0.2151
& 0.1924 & 0.1913 & \textbf{0.2968} \\

& NIMA$\uparrow$
& 4.1549 & \underline{4.4654} & 4.3265 & 4.1146
& 4.3122 & 4.0543 & 4.3491
& 4.4110 & \textbf{4.4681} & 4.4254 \\

& CLIP-IQA$\uparrow$
& 0.2946 & \underline{0.3384} & 0.3145 & 0.3098
& 0.3031 & 0.2721 & 0.3368
& 0.3146 & 0.2516 & \textbf{0.3756} \\

& NIQE$\downarrow$
& 7.3558 & 8.2236 & 7.6405 & 7.8160
& 7.4392 & 7.7840 & \underline{6.8792}
& 7.8140 & 8.0146 & \textbf{6.5142} \\

& PSNR$\uparrow$
& 26.0103 & 25.4452 & \underline{26.9876} & 26.0642
& \textbf{27.5100} & 25.6027 & 25.3962
& 26.0933 & 25.1228 & 26.1280 \\

& SSIM$\uparrow$
& 0.8211 & 0.8143 & \underline{0.8449} & 0.8228
& \textbf{0.8522} & 0.8101 & 0.8083
& 0.8269 & 0.8069 & 0.8269 \\

& LPIPS$\downarrow$
& 0.2482 & 0.2506 & 0.2121 & 0.2582
& \underline{0.1911} & 0.2816 & 0.1942
& 0.2313 & 0.2565 & \textbf{0.1193} \\

\midrule

\multirow{10}{*}{\textbf{RealBlur}}
& CLIP-IQA+$\uparrow$
& 0.3468 & \underline{0.3643} & 0.3451 & 0.3517
& 0.3234 & 0.3058 & 0.3047
& 0.3404 & 0.2885 & \textbf{0.4325} \\

& Q-Align$\uparrow$
& 2.0008 & 1.8426 & 2.0757 & 1.8931
& \underline{2.1174} & 1.9523 & 1.9627
& 2.0184 & 1.9942 & \textbf{2.6699} \\

& MUSIQ$\uparrow$
& 28.1524 & \underline{34.5903} & 31.7451 & 26.1142
& 34.2600 & 28.0733 & 31.9886
& 30.5362 & 33.9946 & \textbf{54.6416} \\

& MANIQA$\uparrow$
& 0.1535 & \underline{0.1998} & 0.1677 & 0.1453
& 0.1712 & 0.1507 & 0.1674
& 0.1573 & 0.1651 & \textbf{0.2792} \\

& NIMA$\uparrow$
& 3.4451 & \underline{3.7263} & 3.4871 & 3.4208
& 3.5445 & 3.5096 & 3.6018
& 3.5406 & 3.6223 & \textbf{3.9406} \\

& CLIP-IQA$\uparrow$
& 0.1862 & \textbf{0.2422} & 0.1831 & 0.2032
& 0.1625 & 0.1530 & 0.1625
& 0.1898 & 0.1656 & \underline{0.2080} \\

& NIQE$\downarrow$
& 6.9172 & 7.7569 & 6.7011 & 7.5244
& \underline{6.2897} & 7.5837 & 6.6155
& 6.9841 & 6.5720 & \textbf{5.8381} \\

& PSNR$\uparrow$
& 22.8488 & 18.7801 & \underline{23.0146} & 22.6906
& \textbf{23.1651} & 22.9199 & 20.7273
& 22.8898 & 22.8734 & 22.8191 \\

& SSIM$\uparrow$
& 0.7312 & 0.6582 & \underline{0.7420} & 0.7251
& \textbf{0.7453} & 0.7278 & 0.6464
& 0.7361 & 0.7278 & 0.7283 \\

& LPIPS$\downarrow$
& 0.3512 & 0.4397 & 0.3341 & 0.3891
& \underline{0.2932} & 0.3953 & 0.3800
& 0.3325 & 0.3081 & \textbf{0.2394} \\

\midrule

\multirow{7}{*}{\textbf{Magic8-Pro}}
& CLIP-IQA+$\uparrow$
& 0.4059 & 0.4099 & 0.4241 & 0.3742
& 0.4293 & 0.3682 & \underline{0.4507}
& 0.4136 & 0.3854 & \textbf{0.4649} \\

& Q-Align$\uparrow$
& 2.2442 & 2.3016 & 2.4204 & 2.1415
& 2.4995 & 2.0522 & \underline{2.5173}
& 2.3510 & 2.2456 & \textbf{2.7296} \\

& MUSIQ$\uparrow$
& 36.9596 & 39.3934 & 40.8948 & 34.2868
& 43.0311 & 31.9925 & \underline{43.7125}
& 40.0396 & 40.4342 & \textbf{49.0853} \\

& MANIQA$\uparrow$
& 0.2596 & 0.2691 & 0.2825 & 0.2561
& \underline{0.2876} & 0.2436 & 0.2789
& 0.2654 & 0.2793 & \textbf{0.3226} \\

& NIMA$\uparrow$
& 4.0987 & 4.2333 & 4.1593 & 4.0518
& 4.1929 & 4.0194 & 4.2100
& \underline{4.2348} & 4.2033 & \textbf{4.3350} \\

& CLIP-IQA$\uparrow$
& 0.4003 & \textbf{0.4535} & 0.4220 & 0.4024
& 0.3662 & 0.3713 & 0.3897
& \underline{0.4313} & 0.3295 & 0.4115 \\

& NIQE$\downarrow$
& 9.2072 & 9.5287 & \underline{8.6884} & 10.0782
& 8.7316 & 10.5825 & \textbf{8.1148}
& 9.4899 & 9.6200 & 8.7020 \\

\bottomrule
\end{tabular}%
}
\end{table*}

\subsubsection{Implementation Details}
\label{sec:implementation_details}

\paragraph{Training Details.}
RawMoment uses the pretrained RGB VAE and DiT from
Qwen-Image-2.1~\citep{qwen2026qwenimage21}.
We train in two stages, each for $10^5$ AdamW updates with
random $512\times512$ crops and a global batch size of 4.
In Stage~1, we optimize only the RAW
encoder $E_{\Phi}$ with \cref{eq:alignment_objective}, using
a learning rate of $2\times10^{-5}$ and zero weight decay.
In Stage~2, we freeze $E_{\Phi}$ and jointly optimize the DiT
attention LoRA adapters, RSE, and SPM with
\cref{eq:flow_matching}, using a learning rate of $10^{-4}$
and weight decay of 0.01. All experiments are conducted on a single NVIDIA B200 GPU.

\paragraph{Compared Methods.}
We compare our RawMoment with nine representative baselines,
categorized into two groups.
The traditional methods include SID, RAWNet~\citep{liang2020raw}, DeepRFT~\citep{mao2023intriguing},
DNF~\citep{jin2023dnf}, LoFormer~\citep{mao2024loformer}, and RAWMamba~\citep{chen2025retinex},
while the generative methods include LDM-ISP~\citep{wen2025ldm},
ISPDiffuser~\citep{ren2025ispdiffuser}, and MeanFlow-RAW~\citep{wang2026extreme}.

\paragraph{Evaluation metrics.}
We employ both no-reference and full-reference metrics
to evaluate image quality.
\textit{No-reference metrics.}
On all three benchmarks, we report seven metrics:
CLIP-IQA+~\citep{wang2023exploring}, Q-Align~\citep{wu2023q}, MUSIQ~\citep{ke2021musiq}, MANIQA~\citep{yang2022maniqa}, NIMA~\citep{talebi2018nima}, CLIP-IQA~\citep{wang2023exploring},
and NIQE~\citep{mittal2012making}.
\textit{Full-reference metrics.}
On RawMoment-Bench and RealBlur, where reference images
are available, we additionally report PSNR, SSIM~\citep{wang2004image},
and LPIPS~\citep{zhang2018unreasonable}.

\subsection{Experimental Results}
\label{sec:experimental_results}

\paragraph{Quantitative comparison.}
\label{sec:quantitative_comparison}
As shown in \cref{tab:blur_iqa}, RawMoment leads 17 of
21 no-reference comparisons across three datasets.
On RawMoment-Bench, it reduces LPIPS by 37.6\%
compared with the best competing method.
On RealBlur, it reduces LPIPS from 0.2932 to 0.2394.
Trained on synthetic pairs, RawMoment generalizes
to real RAW captures without camera-specific fine-tuning.

\begin{figure*}[t]
    \centering
    \includegraphics[width=\linewidth]{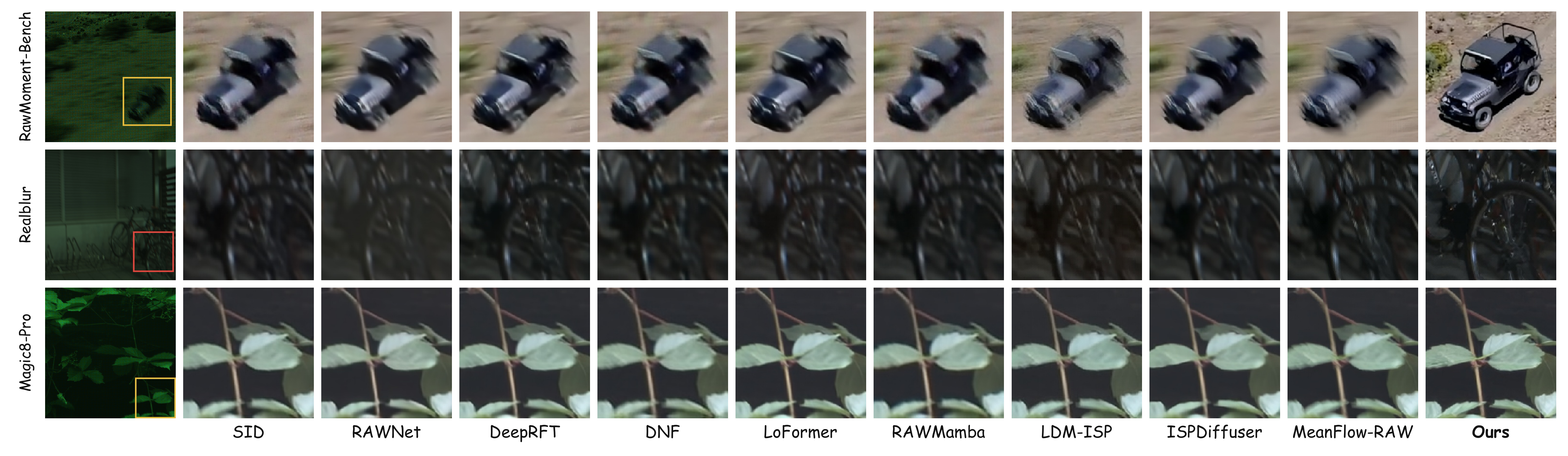}
    \caption{Visual comparison on RawMoment-Bench, RealBlur, and
    Magic8-Pro. The left column shows RAW inputs with highlighted
    crop regions; the remaining columns show the corresponding
    restored crops.}
    \label{fig:visual_comparison}
\end{figure*}

\paragraph{Visual comparison.}
\label{sec:visual_comparison}
As shown in \cref{fig:visual_comparison}, RawMoment
recovers sharper vehicle contours on RawMoment-Bench,
clearer bicycle spokes on RealBlur, and finer leaf
details on Magic8-Pro. These results highlight RawMoment's ability to recover fine details across diverse scenes and cameras.

\begin{figure}[!t]
    \centering
    \includegraphics[width=\linewidth]{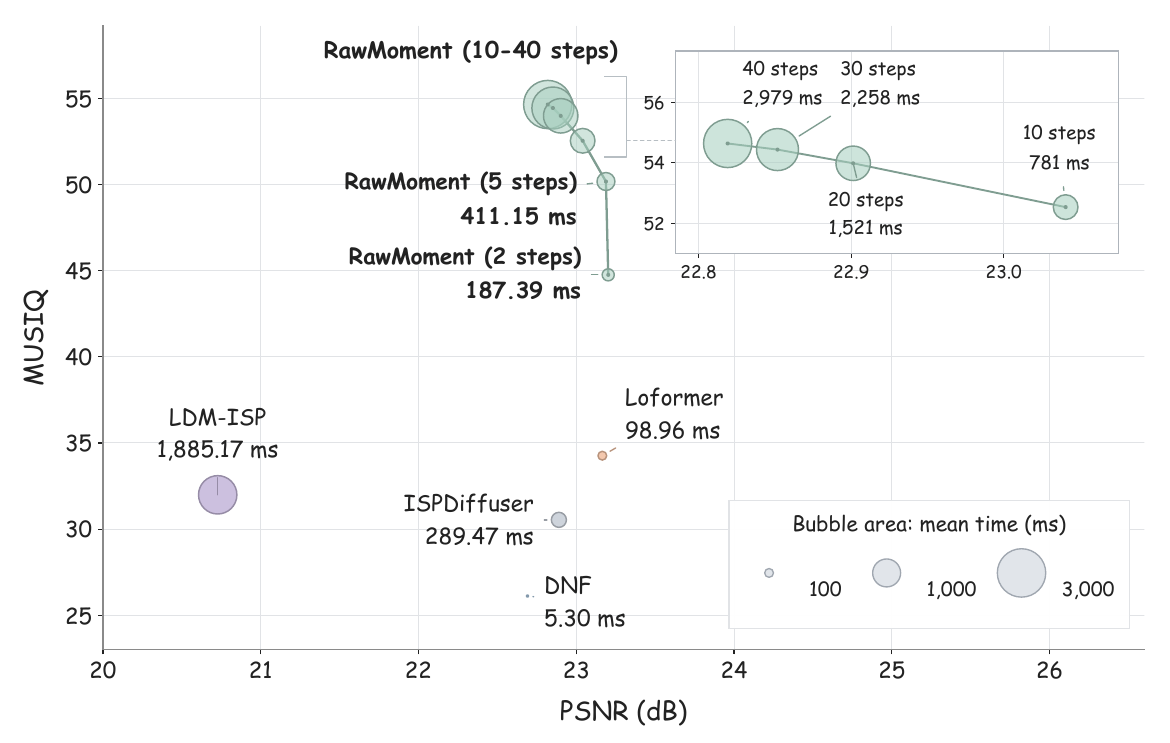}
    \caption{Quality and inference-time comparison on RealBlur.}
    \label{fig:efficiency_comparison}
\end{figure}

\paragraph{Efficiency comparison.}
\label{sec:efficiency_comparison}
We investigate whether RawMoment can reduce inference costs
while retaining its perceptual advantage over competing methods.
We compare RawMoment with DNF, LoFormer, ISPDiffuser, and
LDM-ISP on RealBlur, using a single NVIDIA B200 GPU,
$512\times512$ sensor crops, and a batch size of 1.
For RawMoment with LoRA rank 256, we vary the Euler sampling
steps $K\in\{2,5,10,20,30,40\}$, while competing methods
follow their original inference settings.
Inference time is averaged over 980 images after five warm-up
passes, excluding data transfer, I/O, and metric computation.
As shown in \cref{fig:efficiency_comparison}, reducing $K$
from 40 to 5 decreases latency from 2,979.04 to 411.15\,ms,
with PSNR changing from 22.82 to 23.19\,dB and MUSIQ
from 54.64 to 50.17.
Notably, even with fewer sampling steps, RawMoment maintains
its advantage in no-reference image quality over competing
methods while achieving faster inference than LDM-ISP.
Further improving inference efficiency remains an important
direction for future work.

\begin{figure}[t]
    \centering
    \includegraphics[width=\linewidth]{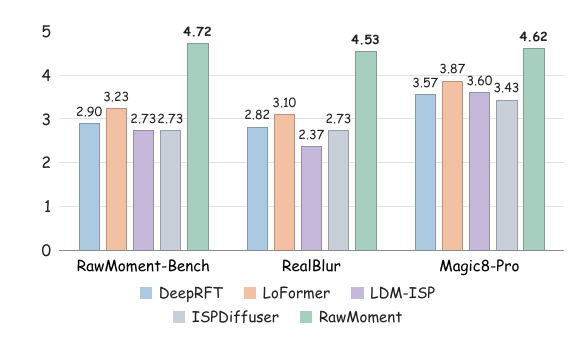}
    \caption{Pilot user study of overall visual quality.}
    \label{fig:user_study}
\end{figure}

\paragraph{User study.}
\label{sec:user_study}
We conduct a pilot user study with three observers from the
field of computational photography, comparing RawMoment with
DeepRFT, LoFormer, LDM-ISP, and ISPDiffuser.
Each observer evaluates 60 images, with 20 from each benchmark,
and rates the overall visual quality of each method on a
five-point scale from 1 (very poor) to 5 (very good).
We aggregate the ratings into mean opinion scores (MOS).
As shown in \cref{fig:user_study}, RawMoment achieves the
highest MOS on all three benchmarks.
Its overall MOS is 4.62, 1.22 points higher than LoFormer.

\subsection{Ablation Studies}
\label{sec:ablation_studies}

\paragraph{Dataset ablation.}
\label{sec:dataset_ablation}
To assess the role of our synthetic data in cross-camera
generalization, we train DNF, LoFormer, ISPDiffuser, LDM-ISP,
and RawMoment on either RawMoment or RealBlur, keeping each
method's architecture and inference protocol fixed.
As shown in \cref{tab:dataset_ablation}, training on RawMoment
improves MUSIQ for all five methods on the external Magic8-Pro
dataset.
DNF rises from 28.07 to 34.29, while ISPDiffuser and RawMoment
gain 5.04 and 0.89 points, respectively.
These consistent gains across restoration models support the
value of our synthetic data for generalization to real RAW
captures.
\begin{table}[t]
\centering
\caption{Dataset ablation on Magic8-Pro.
RB and RM denote training on RealBlur and RawMoment, respectively.
$\Delta$ is RawMoment minus RealBlur, computed before rounding.}
\label{tab:dataset_ablation}
\tablebody

\begin{tabular}{l|ccc|ccc}
\toprule
\multirow{2}{*}{\textbf{Method}}
& \multicolumn{3}{c|}{\textbf{CLIP-IQA+$\uparrow$}}
& \multicolumn{3}{c}{\textbf{MUSIQ$\uparrow$}} \\
\cmidrule(lr){2-4}
\cmidrule(lr){5-7}
& \textbf{RB} & \textbf{RM} & $\Delta\uparrow$
& \textbf{RB} & \textbf{RM} & $\Delta\uparrow$ \\
\midrule
DNF
& 0.3076 & \textbf{0.3742} & $+0.0667$
& 28.0654 & \textbf{34.2868} & $+6.2214$ \\
LoFormer
& 0.4190 & \textbf{0.4293} & $+0.0103$
& 38.8708 & \textbf{43.0311} & $+4.1604$ \\
ISPDiffuser
& 0.3750 & \textbf{0.4136} & $+0.0385$
& 34.9968 & \textbf{40.0396} & $+5.0428$ \\
LDM-ISP
& 0.3539 & \textbf{0.4507} & $+0.0968$
& 37.7333 & \textbf{43.7125} & $+5.9792$ \\
\midrule
\textbf{Ours}
& \textbf{0.4655} & 0.4649 & $-0.0006$
& 48.1996 & \textbf{49.0853} & $+0.8857$ \\
\bottomrule
\end{tabular}
\end{table}

\paragraph{Latent adaptation.}
\label{sec:latent_adaptation_ablation}
\cref{tab:vae_reconstruction} evaluates latent adaptation (LA)
by comparing the RAW encoder at its RGB-pretrained
initialization with CFA input mapping (w/o LA) and after
$10^5$ Stage~1 updates (w/ LA).
Both settings encode identical blurred RAW inputs and
decode posterior means using the same frozen RGB decoder,
without Stage~2.
On RawMoment-Bench, LA improves PSNR from 16.548 to
41.027\,dB and reduces LPIPS from 0.6703 to 0.0370.
Similar improvements are observed on RealBlur, with PSNR
increasing from 11.800 to 25.238\,dB and LPIPS decreasing
from 0.9281 to 0.1303.
These results establish a reliable latent representation
of RAW observations for subsequent joint ISP and deblurring.
\begin{table}[t]
\centering
\caption{VAE reconstruction before (w/o LA) and after (w/ LA)
Stage~1 encoder adaptation on RawMoment-Bench and RealBlur.
Both settings share the frozen RGB decoder and are evaluated
against the corresponding blurred RGB references.
RealBlur metrics use per-image brightness alignment.}
\label{tab:vae_reconstruction}
\tablebody
\begin{tabular}{llccc}
\toprule
\textbf{Dataset} & \textbf{Setting} & \textbf{PSNR$\uparrow$}
& \textbf{SSIM$\uparrow$} & \textbf{LPIPS$\downarrow$} \\
\midrule
\multirow{2}{*}{RawMoment-Bench}
& w/o LA & 16.548 & 0.2964 & 0.6703 \\
& w/ LA & \textbf{41.027} & \textbf{0.9716} & \textbf{0.0370} \\
\midrule
\multirow{2}{*}{RealBlur}
& w/o LA & 11.800 & 0.0750 & 0.9281 \\
& w/ LA & \textbf{25.238} & \textbf{0.8814} & \textbf{0.1303} \\
\bottomrule
\end{tabular}
\end{table}

\paragraph{Module ablation.}
\cref{tab:component_ablation} shows that both SPM and RSE
contribute to restoration quality on RealBlur.
Without SPM, PSNR falls by 0.41\,dB and MUSIQ drops from
54.64 to 51.17.
Removing RSE incurs a larger PSNR loss of 0.57\,dB, along
with a 2.00-point decline in MUSIQ.
Together, the two modules yield gains of 0.36\,dB in PSNR and
5.15 MUSIQ points over the variant with both removed.
These results suggest that sensor-aware physical conditioning
and RAW spatial encoding provide complementary benefits for
joint RAW ISP and deblurring.
\begin{table}[t]
\centering
\caption{Component ablation study on RealBlur.}
\label{tab:component_ablation}

\tablebody

\begin{tabular}{lcc|cc|ccc}
\toprule
\multirow{2}{*}{\textbf{Method}}
& \multirow{2}{*}{\textbf{SPM}}
& \multirow{2}{*}{\textbf{RSE}}
& \multicolumn{2}{c|}{\textbf{No-reference}}
& \multicolumn{3}{c}{\textbf{Full-reference}} \\
\cmidrule(lr){4-5}
\cmidrule(lr){6-8}
& &
& \shortstack{\textbf{CLIP-}\\\textbf{IQA+$\uparrow$}}
& \textbf{MUSIQ$\uparrow$}
& \textbf{PSNR$\uparrow$}
& \textbf{SSIM$\uparrow$}
& \textbf{LPIPS$\downarrow$} \\
\midrule

w/o Both
& $\times$ & $\times$
& 0.3844 & 49.4919
& 22.4599 & 0.7193 & 0.2494 \\

w/o SPM
& $\times$ & $\checkmark$
& 0.3971 & 51.1711
& 22.4056 & 0.7155 & 0.2497 \\

w/o RSE
& $\checkmark$ & $\times$
& 0.4172 & 52.6434
& 22.2465 & 0.7130 & 0.2464 \\

\midrule

\textbf{Ours}
& $\checkmark$ & $\checkmark$
& \textbf{0.4325}
& \textbf{54.6416}
& \textbf{22.8191}
& \textbf{0.7283}
& \textbf{0.2394} \\

\bottomrule
\end{tabular}
\end{table}

\paragraph{Hyperparameter ablation.}
We use LoRA rank 256 by default to balance reconstruction
fidelity, perceptual quality, and adapter size.
As shown in \cref{tab:rank_ablation}, it achieves the highest
PSNR of 22.82\,dB, with MUSIQ rising from 49.76 at rank 32
to 54.64.
Doubling the rank to 512 adds just 0.41 MUSIQ points, but
lowers PSNR by 0.24\,dB and doubles the number of LoRA
parameters.
\begin{table}[t]
\centering
\caption{Ablation study of LoRA ranks on the RealBlur dataset.}
\label{tab:rank_ablation}
\tablebody

\begin{tabular}{l|cc|ccc}
\toprule
\multirow{2}{*}{\textbf{Rank}}
& \multicolumn{2}{c|}{\textbf{No-reference}}
& \multicolumn{3}{c}{\textbf{Full-reference}} \\
\cmidrule(lr){2-3}
\cmidrule(lr){4-6}
&
\textbf{CLIP-IQA+$\uparrow$}
& \textbf{MUSIQ$\uparrow$}
& \textbf{PSNR$\uparrow$}
& \textbf{SSIM$\uparrow$}
& \textbf{LPIPS$\downarrow$} \\
\midrule

32
& 0.3972
& 49.7612
& 22.5565
& 0.7236
& 0.2385 \\

64
& 0.3974
& 50.0729
& 22.4114
& 0.7192
& \textbf{0.2381} \\

128
& 0.4144
& 53.0019
& 22.2295
& 0.7098
& 0.2498 \\

256
& 0.4325
& 54.6416
& \textbf{22.8191}
& \textbf{0.7283}
& 0.2394 \\

512
& \textbf{0.4331}
& \textbf{55.0523}
& 22.5803
& 0.7251
& 0.2420 \\

\bottomrule
\end{tabular}
\end{table}

\subsection{Sensitivity Analysis}
\label{sec:sensitivity_analysis}
To assess the stability of our method under inference randomness,
we evaluate ten different seeds with all other settings unchanged.
\cref{tab:seed_sensitivity} aggregates scores computed on each
complete test set. The PSNR standard deviations are 0.012\,dB on
RealBlur and 0.057\,dB on RawMoment-Bench.
These results indicate stable performance across seeds, while
maintaining state-of-the-art MUSIQ scores among the compared
methods on all three benchmarks.
\begin{table}[t]
\centering
\caption{RawMoment's performance across 10 different random seeds on three datasets.}
\label{tab:seed_sensitivity}
\tablebody
\setlength{\tabcolsep}{1pt}
\begin{tabular}{@{}llcccc@{}}
\toprule
\textbf{Dataset} & \textbf{Metric} & \textbf{First}
& \textbf{Mean $\pm$ Std.} & \textbf{Worst} & \textbf{Best} \\
\midrule
\multirow{2}{*}{\shortstack[l]{RawMoment-\\Bench}} & PSNR & 26.128 & $26.032 \pm 0.057$ & 25.955 & 26.128 \\
 & MUSIQ & 56.088 & $55.936 \pm 0.193$ & 55.685 & 56.248 \\
\midrule
\multirow{2}{*}{RealBlur} & PSNR & 22.819 & $22.821 \pm 0.012$ & 22.806 & 22.844 \\
 & MUSIQ & 54.642 & $54.546 \pm 0.063$ & 54.406 & 54.642 \\
\midrule
\multirow{2}{*}{Magic8-Pro} & MUSIQ & 49.085 & $49.269 \pm 0.240$ & 48.716 & 49.627 \\
 & CLIP-IQA+ & 0.4649 & $0.4618 \pm 0.0018$ & 0.4586 & 0.4649 \\
\bottomrule
\end{tabular}
\end{table}

\section{Conclusion}
\label{sec:conclusion}
In this paper, we presented RawMoment, a generative framework for joint RAW ISP and motion deblurring. To address the scarcity of diverse paired RAW data, we developed a scalable data construction pipeline that combines real-video motion with calibrated sensor synthesis, and established RawMoment-Bench for evaluation. To leverage pretrained text-to-image priors, RawMoment aligns RAW observations with the pretrained RGB latent space and introduces spatial and physical conditioning to guide sharp RGB reconstruction. Extensive experiments demonstrate superior perceptual quality across three benchmarks and strong generalization to real RAW captures from unseen cameras without camera-specific fine-tuning. Ablation studies further validate the effectiveness of our data construction and individual framework components. While this work primarily focuses on restoration quality, improving computational efficiency remains an important direction for future research. 
In particular, we plan to investigate accelerated generative sampling and lightweight model architectures to reduce inference latency and memory consumption, thereby facilitating the practical deployment of high-quality generative RAW-to-RGB deblurring in real-world photography and resource-constrained imaging systems.

\section*{Acknowledgments}
We thank Jianan Sun, Bin Chen, and Yu Wang of Honor Device Co., Ltd.
for their help and support.

\section{LLM Usage Statement}
We used ChatGPT to assist with manuscript polishing, LaTeX formatting, and verification of experimental results. All content and results were reviewed and validated by the authors, who take full responsibility for the accuracy and integrity of this work.

{
    \small
    \bibliographystyle{ieeenat_fullname}
    \bibliography{main,sec/supp_refs}

\begin{thebibliography}{37}
\providecommand{\natexlab}[1]{#1}
\providecommand{\url}[1]{\texttt{#1}}
\expandafter\ifx\csname urlstyle\endcsname\relax
  \providecommand{\doi}[1]{doi: #1}\else
  \providecommand{\doi}{doi: \begingroup \urlstyle{rm}\Url}\fi

\bibitem[Bai et~al.(2025)Bai, Cai, Chen, Chen, Chen, Cheng, Deng, Ding, Gao, Ge, Ge, Guo, Huang, Huang, Huang, Hui, Jiang, Li, Li, Li, Li, Lin, Lin, Liu, Liu, Liu, Liu, Liu, Liu, Lu, Luo, Lv, Men, Meng, Ren, Ren, Song, Sun, Tang, Tu, Wan, Wang, Wang, Wang, Wang, Xie, Xu, Xu, Xu, Yang, Yang, Yang, Yang, Yu, Zhang, Zhang, Zhang, Zheng, Zhong, Zhou, Zhou, Zhou, Zhu, and Zhu]{bai2025qwen3vl}
Shuai Bai, Yuxuan Cai, Ruizhe Chen, Keqin Chen, Xionghui Chen, Zesen Cheng, Lianghao Deng, Wei Ding, Chang Gao, Chunjiang Ge, Wenbin Ge, Zhifang Guo, Qidong Huang, Jie Huang, Fei Huang, Binyuan Hui, Shutong Jiang, Zhaohai Li, Mingsheng Li, Mei Li, Kaixin Li, Zicheng Lin, Junyang Lin, Xuejing Liu, Jiawei Liu, Chenglong Liu, Yang Liu, Dayiheng Liu, Shixuan Liu, Dunjie Lu, Ruilin Luo, Chenxu Lv, Rui Men, Lingchen Meng, Xuancheng Ren, Xingzhang Ren, Sibo Song, Yuchong Sun, Jun Tang, Jianhong Tu, Jianqiang Wan, Peng Wang, Pengfei Wang, Qiuyue Wang, Yuxuan Wang, Tianbao Xie, Yiheng Xu, Haiyang Xu, Jin Xu, Zhibo Yang, Mingkun Yang, Jianxin Yang, An Yang, Bowen Yu, Fei Zhang, Hang Zhang, Xi Zhang, Bo Zheng, Humen Zhong, Jingren Zhou, Fan Zhou, Jing Zhou, Yuanzhi Zhu, and Ke Zhu.
\newblock {Qwen3-VL} technical report.
\newblock \emph{arXiv preprint arXiv:2511.21631}, 2025.

\bibitem[Chen et~al.(2018)Chen, Chen, Xu, and Koltun]{chen2018learning}
Chen Chen, Qifeng Chen, Jia Xu, and Vladlen Koltun.
\newblock Learning to see in the dark.
\newblock In \emph{2018 IEEE/CVF Conference on Computer Vision and Pattern Recognition}, pages 3291--3300. Ieee, 2018.

\bibitem[Chen et~al.(2026)Chen, Han, Huang, Feng, Zhang, and Han]{chen2025retinex}
Xianmin Chen, Longfei Han, Peiliang Huang, Xiaoxu Feng, Dingwen Zhang, and Junwei Han.
\newblock Retinex-rawmamba: Bridging demosaicing and denoising for low-light raw image enhancement.
\newblock \emph{IEEE Transactions on Circuits and Systems for Video Technology}, 36\penalty0 (1):\penalty0 406--420, 2026.

\bibitem[Chen et~al.(2025)Chen, Wen, Li, Liu, Guo, Hu, and Chen]{chen2025rddm}
Yan Chen, Yi Wen, Wei Li, Junchao Liu, Yong Guo, Jie Hu, and Xinghao Chen.
\newblock {RDDM}: Practicing {RAW} domain diffusion model for real-world image restoration.
\newblock \emph{arXiv preprint arXiv:2508.19154}, 2025.

\bibitem[Cui and Harada(2024)]{cui2024rawadapter}
Ziteng Cui and Tatsuya Harada.
\newblock {RAW-Adapter}: Adapting pre-trained visual model to camera {RAW} images.
\newblock In \emph{ECCV}, 2024.

\bibitem[Deng et~al.(2025)Deng, Wu, Yang, Zhu, Wang, and Wu]{deng2025acquire}
Junyuan Deng, Xinyi Wu, Yongxing Yang, Congchao Zhu, Song Wang, and Zhenyao Wu.
\newblock Acquire and then adapt: Squeezing out text-to-image model for image restoration.
\newblock In \emph{CVPR}, pages 23195--23206, 2025.

\bibitem[Fei et~al.(2026)Fei, Ye, Wang, and Zhu]{fei2026lucidflux}
Song Fei, Tian Ye, Lujia Wang, and Lei Zhu.
\newblock Lucidflux: Caption-free universal image restoration via a large-scale diffusion transformer.
\newblock In \emph{International Conference on Learning Representations}, pages 86233--86260, 2026.

\bibitem[Friedman and Dieng(2023)]{friedman2023vendi}
Dan Friedman and Adji~Bousso Dieng.
\newblock The {Vendi Score}: A diversity evaluation metric for machine learning.
\newblock \emph{Transactions on Machine Learning Research}, 2023.

\bibitem[Hu et~al.(2022)Hu, Shen, Wallis, Allen-Zhu, Li, Wang, Wang, and Chen]{hu2022lora}
Edward~J. Hu, Yelong Shen, Phillip Wallis, Zeyuan Allen-Zhu, Yuanzhi Li, Shean Wang, Lu Wang, and Weizhu Chen.
\newblock {LoRA}: Low-rank adaptation of large language models.
\newblock In \emph{ICLR}, 2022.

\bibitem[{International Telecommunication Union}(2015)]{itu2015bt709}
{International Telecommunication Union}.
\newblock {Parameter values for the HDTV standards for production and international programme exchange}.
\newblock Recommendation ITU-R BT.709-6, International Telecommunication Union, 2015.

\bibitem[Jiao et~al.(2025)Jiao, Li, Shang, Wang, and Ren]{jiao2025efficient}
Wenlong Jiao, Binglong Li, Wei Shang, Ping Wang, and Dongwei Ren.
\newblock Efficient {RAW} image deblurring with adaptive frequency modulation.
\newblock In \emph{NeurIPS}, 2025.

\bibitem[Jin et~al.(2023)Jin, Han, Li, Guo, Chai, and Li]{jin2023dnf}
Xin Jin, Ling-Hao Han, Zhen Li, Chun-Le Guo, Zhi Chai, and Chongyi Li.
\newblock Dnf: Decouple and feedback network for seeing in the dark.
\newblock In \emph{2023 IEEE/CVF Conference on Computer Vision and Pattern Recognition (CVPR)}, pages 18135--18144. IEEE, 2023.

\bibitem[Ke et~al.(2021)Ke, Wang, Wang, Milanfar, and Yang]{ke2021musiq}
Junjie Ke, Qifei Wang, Yilin Wang, Peyman Milanfar, and Feng Yang.
\newblock Musiq: Multi-scale image quality transformer.
\newblock In \emph{2021 IEEE/CVF International Conference on Computer Vision (ICCV)}, pages 5128--5137. IEEE, 2021.

\bibitem[Liang et~al.(2022)Liang, Chen, Liu, and Hsu]{liang2020raw}
Chih-Hung Liang, Yu-An Chen, Yueh-Cheng Liu, and Winston~H Hsu.
\newblock {RAW} image deblurring.
\newblock \emph{IEEE Transactions on Multimedia}, 24:\penalty0 61--72, 2022.

\bibitem[Lipman et~al.(2023)Lipman, Chen, Ben-Hamu, Nickel, and Le]{lipman2023flow}
Yaron Lipman, Ricky T.~Q. Chen, Heli Ben-Hamu, Maximilian Nickel, and Matt Le.
\newblock Flow matching for generative modeling.
\newblock In \emph{ICLR}, 2023.

\bibitem[Mao et~al.(2023)Mao, Liu, Liu, Li, Shen, and Wang]{mao2023intriguing}
Xintian Mao, Yiming Liu, Fengze Liu, Qingli Li, Wei Shen, and Yan Wang.
\newblock Intriguing findings of frequency selection for image deblurring.
\newblock In \emph{Proceedings of the AAAI conference on artificial intelligence}, pages 1905--1913, 2023.

\bibitem[Mao et~al.(2024)Mao, Wang, Xie, Li, and Wang]{mao2024loformer}
Xintian Mao, Jiansheng Wang, Xingran Xie, Qingli Li, and Yan Wang.
\newblock Loformer: Local frequency transformer for image deblurring.
\newblock In \emph{Proceedings of the 32nd ACM international conference on multimedia}, pages 10382--10391, 2024.

\bibitem[Mittal et~al.(2013)Mittal, Soundararajan, and Bovik]{mittal2012making}
Anish Mittal, Rajiv Soundararajan, and Alan~C Bovik.
\newblock Making a “completely blind” image quality analyzer.
\newblock \emph{IEEE Signal processing letters}, 20\penalty0 (3):\penalty0 209--212, 2013.

\bibitem[Nah et~al.(2017)Nah, Kim, and Lee]{nah2017deep}
Seungjun Nah, Tae~Hyun Kim, and Kyoung~Mu Lee.
\newblock Deep multi-scale convolutional neural network for dynamic scene deblurring.
\newblock In \emph{CVPR}, pages 3883--3891, 2017.

\bibitem[Nan et~al.(2025)Nan, Xie, Zhou, Fan, Yang, Chen, Li, Yang, and Tai]{nan2025openvid}
Kepan Nan, Rui Xie, Penghao Zhou, Tiehan Fan, Zhenheng Yang, Zhijie Chen, Xiang Li, Jian Yang, and Ying Tai.
\newblock {OpenVid-1M}: A large-scale high-quality dataset for text-to-video generation.
\newblock In \emph{International Conference on Learning Representations}, 2025.

\bibitem[Oquab et~al.(2024)Oquab, Darcet, Moutakanni, Vo, Szafraniec, Khalidov, Fernandez, Haziza, Massa, El-Nouby, Assran, Ballas, Galuba, Howes, Huang, Li, Misra, Rabbat, Sharma, Synnaeve, Xu, J{\'e}gou, Mairal, Labatut, Joulin, and Bojanowski]{oquab2024dinov2}
Maxime Oquab, Timoth{\'e}e Darcet, Th{\'e}o Moutakanni, Huy~V. Vo, Marc Szafraniec, Vasil Khalidov, Pierre Fernandez, Daniel Haziza, Francisco Massa, Alaaeldin El-Nouby, Mahmoud Assran, Nicolas Ballas, Wojciech Galuba, Russell Howes, Po-Yao Huang, Shang-Wen Li, Ishan Misra, Michael Rabbat, Vasu Sharma, Gabriel Synnaeve, Hu Xu, Herv{\'e} J{\'e}gou, Julien Mairal, Patrick Labatut, Armand Joulin, and Piotr Bojanowski.
\newblock {DINOv2}: Learning robust visual features without supervision.
\newblock \emph{Transactions on Machine Learning Research}, 2024.

\bibitem[{Qwen}(2026)]{qwen2026qwenimage21}
{Qwen}.
\newblock {Qwen-Image-2.1}.
\newblock \url{https://qwen.ai/blog?id=qwen-image-2.1}, 2026.
\newblock Official model release.

\bibitem[Ren et~al.(2025)Ren, Jiang, Yang, Li, and Liu]{ren2025ispdiffuser}
Yang Ren, Hai Jiang, Menglong Yang, Wei Li, and Shuaicheng Liu.
\newblock Ispdiffuser: Learning raw-to-srgb mappings with texture-aware diffusion models and histogram-guided color consistency.
\newblock In \emph{Proceedings of the AAAI Conference on Artificial Intelligence}, pages 6722--6730, 2025.

\bibitem[Rim et~al.(2020)Rim, Lee, Won, and Cho]{rim2020realblur}
Jaesung Rim, Haeyun Lee, Jucheol Won, and Sunghyun Cho.
\newblock Real-world blur dataset for learning and benchmarking deblurring algorithms.
\newblock In \emph{ECCV}, 2020.

\bibitem[Rim et~al.(2022)Rim, Kim, Kim, Lee, Lee, and Cho]{rim2022realistic}
Jaesung Rim, Geonung Kim, Jungeon Kim, Junyong Lee, Seungyong Lee, and Sunghyun Cho.
\newblock Realistic blur synthesis for learning image deblurring.
\newblock In \emph{ECCV}, 2022.

\bibitem[Sun et~al.(2025)Sun, Wu, Ma, Liu, Yi, and Zhang]{sun2025pixel}
Lingchen Sun, Rongyuan Wu, Zhiyuan Ma, Shuaizheng Liu, Qiaosi Yi, and Lei Zhang.
\newblock Pixel-level and semantic-level adjustable super-resolution: A dual-{LoRA} approach.
\newblock In \emph{Proceedings of the IEEE/CVF Conference on Computer Vision and Pattern Recognition}, pages 2333--2343, 2025.

\bibitem[Talebi and Milanfar(2018)]{talebi2018nima}
Hossein Talebi and Peyman Milanfar.
\newblock Nima: Neural image assessment.
\newblock \emph{IEEE transactions on image processing}, 27\penalty0 (8):\penalty0 3998--4011, 2018.

\bibitem[van~der Maaten and Hinton(2008)]{vandermaaten2008tsne}
Laurens van~der Maaten and Geoffrey Hinton.
\newblock Visualizing data using {t-SNE}.
\newblock \emph{Journal of Machine Learning Research}, 9\penalty0 (86):\penalty0 2579--2605, 2008.

\bibitem[Wang et~al.(2023)Wang, Chan, and Loy]{wang2023exploring}
Jianyi Wang, Kelvin~CK Chan, and Chen~Change Loy.
\newblock Exploring clip for assessing the look and feel of images.
\newblock In \emph{Proceedings of the AAAI conference on artificial intelligence}, pages 2555--2563, 2023.

\bibitem[Wang et~al.(2024)Wang, Lipson, and Deng]{wang2024searaft}
Yihan Wang, Lahav Lipson, and Jia Deng.
\newblock Sea-raft: Simple, efficient, accurate raft for optical flow.
\newblock In \emph{European Conference on Computer Vision}, pages 36--54. Springer, 2024.

\bibitem[Wang et~al.(2004)Wang, Bovik, Sheikh, and Simoncelli]{wang2004image}
Zhou Wang, Alan~C Bovik, Hamid~R Sheikh, and Eero~P Simoncelli.
\newblock Image quality assessment: from error visibility to structural similarity.
\newblock \emph{IEEE transactions on image processing}, 13\penalty0 (4):\penalty0 600--612, 2004.

\bibitem[Wang et~al.(2026)Wang, Liang, Xiao, and Chen]{wang2026extreme}
Zepu Wang, Jingze Liang, Weijie Xiao, and Kexin Chen.
\newblock When extreme darkness meets motion blur: Meanflow for unified raw restoration.
\newblock \emph{arXiv preprint arXiv:2608.01720}, 2026.

\bibitem[Wen et~al.(2025)Wen, Rao, Xing, and Chen]{wen2025ldm}
Qiang Wen, Zhefan Rao, Yazhou Xing, and Qifeng Chen.
\newblock Ldm-isp: Enhancing neural isp for low light with latent diffusion models.
\newblock In \emph{2025 IEEE International Conference on Robotics and Automation (ICRA)}, pages 13297--13303. IEEE, 2025.

\bibitem[Wu et~al.(2024)Wu, Zhang, Zhang, Chen, Liao, Li, Gao, Wang, Zhang, Sun, Yan, Min, Zhai, and Lin]{wu2023q}
Haoning Wu, Zicheng Zhang, Weixia Zhang, Chaofeng Chen, Liang Liao, Chunyi Li, Yixuan Gao, Annan Wang, Erli Zhang, Wenxiu Sun, Qiong Yan, Xiongkuo Min, Guangtao Zhai, and Weisi Lin.
\newblock Q-align: Teaching {LMM}s for visual scoring via discrete text-defined levels.
\newblock In \emph{Proceedings of the 41st International Conference on Machine Learning}, pages 54015--54029. PMLR, 2024.

\bibitem[Yang et~al.(2022)Yang, Wu, Shi, Lao, Gong, Cao, Wang, and Yang]{yang2022maniqa}
Sidi Yang, Tianhe Wu, Shuwei Shi, Shanshan Lao, Yuan Gong, Mingdeng Cao, Jiahao Wang, and Yujiu Yang.
\newblock Maniqa: Multi-dimension attention network for no-reference image quality assessment.
\newblock In \emph{2022 IEEE/CVF Conference on Computer Vision and Pattern Recognition Workshops (CVPRW)}, pages 1190--1199. IEEE, 2022.

\bibitem[Yu et~al.(2024)Yu, Gu, Li, Hu, Kong, Wang, He, Qiao, and Dong]{yu2024scaling}
Fanghua Yu, Jinjin Gu, Zheyuan Li, Jinfan Hu, Xiangtao Kong, Xintao Wang, Jingwen He, Yu Qiao, and Chao Dong.
\newblock Scaling up to excellence: Practicing model scaling for photo-realistic image restoration in the wild.
\newblock In \emph{Proceedings of the IEEE/CVF Conference on Computer Vision and Pattern Recognition}, pages 25669--25680, 2024.

\bibitem[Zhang et~al.(2018)Zhang, Isola, Efros, Shechtman, and Wang]{zhang2018unreasonable}
Richard Zhang, Phillip Isola, Alexei~A Efros, Eli Shechtman, and Oliver Wang.
\newblock The unreasonable effectiveness of deep features as a perceptual metric.
\newblock In \emph{2018 IEEE/CVF conference on computer vision and pattern recognition}, pages 586--595. IEEE, 2018.

\end{thebibliography}
}

\clearpage
\appendix
\setcounter{equation}{0}
\renewcommand{\theequation}{S\arabic{equation}}
\renewcommand{\theHequation}{supp.\arabic{equation}}
\setcounter{table}{0}
\renewcommand{\thetable}{S\arabic{table}}
\renewcommand{\theHtable}{supp.\arabic{table}}
\maketitlesupplementary
\ifdefined\resetlinenumber\resetlinenumber[1]\fi

\section{Sensor RAW Synthesis}
\label{sec:supp_raw_synthesis}
We specify the rendering parameters used for RawMoment and
RawMoment-Bench, following the synthesis pipeline in \cref{sec:data}.

\paragraph{Brightness and white balance.}
We normalize the RGB crops to $[0,1]$ and apply the brightness
transform $y^{\gamma}$ with $\gamma=1.6$.
The normalized black and white levels are $\beta=1/16$ and 1,
respectively.
For each RAW variant, we independently sample the red and blue
white-balance gains as
\begin{equation}
g_R\sim\mathcal{U}[1.9,2.4],
\qquad
g_B\sim\mathcal{U}[1.5,1.9].
\label{eq:supp_white_balance}
\end{equation}
The inverse white-balance vector
$\mathbf{w}=(g_R^{-1},1,g_B^{-1})$ maps the signal to the
domain in which sensor noise is added:
$u=\mathbf{w}\odot(1-\beta)y^{\gamma}$.

\paragraph{Sensor noise.}
In \cref{eq:raw_noise}, $a$ is the shot-noise gain and $b$ is the
read-noise variance, expressed in the normalized signal domain
before white-balance amplification.
The Gaussian noise is generated using independent standard-normal
draws across pixels and color channels, with variance $a\,u+b$.

\paragraph{Pairing and RAW representation.}
Each training RGB pair yields three RAW variants with independently
sampled acquisition settings and noise seeds.
Within a variant, the sharp and blurred renderings share the
acquisition settings and standard-normal draws, while their noise
amplitudes depend on their respective signal intensities.
Thus, the sharp RAW used in Stage~1 retains the low-light transform
and sensor noise.
RawMoment-Bench uses one fixed variant per scene.
The noisy signal is mapped to an RGGB mosaic following
\cref{eq:raw_noise,eq:raw_mosaic} and rounded to unsigned 16-bit integers as
$q=\operatorname{round}(65535\,r)$ and stored in a sparse RGB canvas.
At valid CFA sites, the data loader computes
\begin{equation}
x=\frac{q/65535-\beta}{1-\beta}.
\label{eq:supp_raw_normalization}
\end{equation}
The normalized samples are separated into four R/G1/G2/B channels
at their original CFA positions.
For network input, valid samples are scaled to $2x-1$ and
unsampled entries are zero.

\newpage
\section{Conditioning Details}
\label{sec:supp_conditioning}

\paragraph{RSE configuration.}
For a $512\times512$ input, RSE produces a 64-channel
$32\times32$ feature grid.
Each $2\times2$ neighborhood is concatenated into a 256-dimensional
vector, then mapped by LayerNorm and a linear projection to
4096 dimensions.
Adding the learned RAW-stream embedding yields 256 tokens on a
$16\times16$ grid.

\paragraph{SPM parameter normalization.}
For synthetic data, SPM reads ISO, the noise coefficients $(a,b)$,
and the inverse white-balance vector $\mathbf{w}$ from the
corresponding rendering record.
We construct the eight-dimensional input as
\begin{equation}
\begin{aligned}
\mathbf{p}=\Big[&
\frac{\log_2(\mathrm{ISO}/100)}{8},
\frac{\log_{10}(a)}{12},
\frac{\log_{10}(b+10^{-12})}{12},\\
&\frac{\log_2(w_R)}{4},
\frac{\log_2(w_G)}{4},
\frac{\log_2(w_B)}{4},\beta,1
\Big]^{\top}.
\end{aligned}
\label{eq:physical_parameters}
\end{equation}
The final two entries encode the black and white levels of the
training RAW convention, $\beta=1/16$ and 1.

\paragraph{SPM architecture and injection.}
The MLP $P_{\omega}$ maps the eight inputs through a 128-unit
hidden layer with SiLU activation to $64\cdot4096$ output values:
\begin{equation}
\begin{aligned}
\mathbf{h}&=\operatorname{SiLU}(W_1\mathbf{p}+\mathbf{b}_1),\\
C_{\mathrm{p}}&=\operatorname{reshape}_{64\times4096}
  (W_2\mathbf{h}+\mathbf{b}_2).
\end{aligned}
\label{eq:supp_spm_mlp}
\end{equation}
The 64 physical tokens are appended to the cached text embeddings
and passed through the pretrained text projection $\Pi_{\mathrm{t}}$
before entering the condition prefix in \cref{eq:condition_sequence}.
SPM is jointly optimized with RSE and the attention LoRA adapters~\citep{hu2022lora}
in Stage~2.
During inference, the acquisition parameters and the resulting
physical tokens remain fixed throughout the Euler updates.

\section{Training Data Comparison}
\label{sec:supp_training_data}
\begin{table*}[!t]
\centering
\caption{Training-data comparison. For each method, models are trained separately on the GoPro (GP) and our RawMoment (RM) training sets, then evaluated on four test sets: GoPro, RealBlur, RawMoment-Bench, and Magic8-Pro. $\Delta=\mathrm{RM}-\mathrm{GP}$ is computed before rounding. Bold marks the better training source within each method and metric. Blue $\Delta$ cells favor RM; orange cells favor GP. Darker cells indicate larger changes on a shared scale per metric.}
\label{tab:gopro_dataset_ablation_benchmark}
\begingroup
\fontsize{7}{9}\selectfont
\setlength{\tabcolsep}{0pt}
\renewcommand{\arraystretch}{1.32}
\newlength{\suppAblationDeltaWidth}
\setlength{\suppAblationDeltaWidth}{\dimexpr(\textwidth-354pt)/5\relax}
\begin{tabular}{@{}p{42pt}@{\hspace{5pt}}p{38pt}@{\hspace{3pt}}>{\raggedleft\arraybackslash}p{23pt}@{\hspace{2pt}}>{\raggedleft\arraybackslash}p{23pt}@{\hspace{2pt}}>{\raggedleft\arraybackslash}p{\suppAblationDeltaWidth}@{\hspace{4pt}}>{\raggedleft\arraybackslash}p{23pt}@{\hspace{2pt}}>{\raggedleft\arraybackslash}p{23pt}@{\hspace{2pt}}>{\raggedleft\arraybackslash}p{\suppAblationDeltaWidth}@{\hspace{4pt}}>{\raggedleft\arraybackslash}p{23pt}@{\hspace{2pt}}>{\raggedleft\arraybackslash}p{23pt}@{\hspace{2pt}}>{\raggedleft\arraybackslash}p{\suppAblationDeltaWidth}@{\hspace{4pt}}>{\raggedleft\arraybackslash}p{23pt}@{\hspace{2pt}}>{\raggedleft\arraybackslash}p{23pt}@{\hspace{2pt}}>{\raggedleft\arraybackslash}p{\suppAblationDeltaWidth}@{\hspace{4pt}}>{\raggedleft\arraybackslash}p{23pt}@{\hspace{2pt}}>{\raggedleft\arraybackslash}p{23pt}@{\hspace{2pt}}>{\raggedleft\arraybackslash}p{\suppAblationDeltaWidth}@{}}
\toprule
\multirow{3}{*}{\textbf{Benchmark}} & \multirow{3}{*}{\textbf{Metric}} & \multicolumn{6}{c}{\cellcolor{black!5}\textbf{Traditional Methods}} & \multicolumn{9}{c}{\cellcolor{black!5}\textbf{Generative Methods}} \\
\cmidrule(lr){3-8}\cmidrule(lr){9-17}
 & & \multicolumn{3}{c}{\textbf{DNF}} & \multicolumn{3}{c}{\textbf{LoFormer}} & \multicolumn{3}{c}{\textbf{LDM-ISP}} & \multicolumn{3}{c}{\textbf{ISPDiffuser}} & \multicolumn{3}{c}{\textbf{Ours}} \\
\cmidrule(lr){3-5} \cmidrule(lr){6-8} \cmidrule(lr){9-11} \cmidrule(lr){12-14} \cmidrule(lr){15-17}
 & & GP & RM & $\Delta$ & GP & RM & $\Delta$ & GP & RM & $\Delta$ & GP & RM & $\Delta$ & GP & RM & $\Delta$ \\
\midrule
\multirow{5}{*}{\textbf{GoPro}} & PSNR$\uparrow$ & 25.7532 & \textbf{26.2937} & \cellcolor[HTML]{E8F1FA}\mbox{+0.5404} & 26.6609 & \textbf{28.7529} & \cellcolor[HTML]{A7C7ED}\mbox{+2.0919} & 25.1341 & \textbf{25.6054} & \cellcolor[HTML]{EBF2FB}\mbox{+0.4713} & 25.7163 & \textbf{26.3572} & \cellcolor[HTML]{E4EEFA}\mbox{+0.6408} & \textbf{27.3837} & 27.1095 & \cellcolor[HTML]{FEF7F2}\mbox{$-$0.2742} \\
 & SSIM$\uparrow$ & 0.7902 & \textbf{0.7984} & \cellcolor[HTML]{F4F8FD}\mbox{+0.0083} & 0.8066 & \textbf{0.8610} & \cellcolor[HTML]{B8D2F1}\mbox{+0.0544} & 0.7613 & \textbf{0.7763} & \cellcolor[HTML]{ECF3FB}\mbox{+0.0149} & 0.7873 & \textbf{0.8017} & \cellcolor[HTML]{ECF3FB}\mbox{+0.0144} & \textbf{0.8229} & 0.8192 & \cellcolor[HTML]{FEFCFA}\mbox{$-$0.0037} \\
 & LPIPS$\downarrow$ & \textbf{0.3046} & 0.3123 & \cellcolor[HTML]{FEFCF9}\mbox{+0.0077} & 0.2233 & \textbf{0.2112} & \cellcolor[HTML]{F7FAFD}\mbox{$-$0.0121} & 0.2317 & \textbf{0.2058} & \cellcolor[HTML]{EDF4FB}\mbox{$-$0.0259} & \textbf{0.2643} & 0.2816 & \cellcolor[HTML]{FEF7F2}\mbox{+0.0173} & \textbf{0.1227} & 0.1387 & \cellcolor[HTML]{FEF8F3}\mbox{+0.0160} \\
 & CLIP-IQA+$\uparrow$ & \textbf{0.3118} & 0.2822 & \cellcolor[HTML]{FCF0E5}\mbox{$-$0.0296} & \textbf{0.3485} & 0.3409 & \cellcolor[HTML]{FEFBF8}\mbox{$-$0.0076} & 0.2991 & \textbf{0.4005} & \cellcolor[HTML]{B0CDEF}\mbox{+0.1014} & \textbf{0.2885} & 0.2882 & \cellcolor[HTML]{FFFFFF}\mbox{$-$0.0002} & 0.3857 & \textbf{0.4177} & \cellcolor[HTML]{E6EFFA}\mbox{+0.0320} \\
 & MUSIQ$\uparrow$ & \textbf{29.1788} & 26.2170 & \cellcolor[HTML]{FCEFE3}\mbox{$-$2.9618} & \textbf{37.4959} & 35.5961 & \cellcolor[HTML]{FDF5ED}\mbox{$-$1.8999} & 28.7784 & \textbf{37.4375} & \cellcolor[HTML]{B7D1F0}\mbox{+8.6591} & \textbf{32.5020} & 30.7486 & \cellcolor[HTML]{FDF5EF}\mbox{$-$1.7534} & 44.7750 & \textbf{47.3794} & \cellcolor[HTML]{E9F1FB}\mbox{+2.6044} \\
\midrule
\multirow{5}{*}{\textbf{RealBlur}} & PSNR$\uparrow$ & 21.6438 & \textbf{22.6906} & \cellcolor[HTML]{D3E3F6}\mbox{+1.0469} & 20.7370 & \textbf{23.1651} & \cellcolor[HTML]{98BEEA}\mbox{+2.4281} & 20.5087 & \textbf{20.7273} & \cellcolor[HTML]{F6F9FD}\mbox{+0.2186} & 21.9494 & \textbf{22.8898} & \cellcolor[HTML]{D7E6F7}\mbox{+0.9404} & 22.7820 & \textbf{22.8191} & \cellcolor[HTML]{FDFEFF}\mbox{+0.0370} \\
 & SSIM$\uparrow$ & 0.6804 & \textbf{0.7251} & \cellcolor[HTML]{C5DAF3}\mbox{+0.0447} & 0.6578 & \textbf{0.7453} & \cellcolor[HTML]{8DB7E8}\mbox{+0.0875} & 0.6379 & \textbf{0.6464} & \cellcolor[HTML]{F4F8FD}\mbox{+0.0085} & 0.6967 & \textbf{0.7361} & \cellcolor[HTML]{CCDFF5}\mbox{+0.0394} & 0.7103 & \textbf{0.7283} & \cellcolor[HTML]{E8F0FA}\mbox{+0.0180} \\
 & LPIPS$\downarrow$ & 0.4415 & \textbf{0.3891} & \cellcolor[HTML]{DCE9F8}\mbox{$-$0.0524} & 0.4619 & \textbf{0.2932} & \cellcolor[HTML]{8DB7E8}\mbox{$-$0.1687} & 0.5030 & \textbf{0.3800} & \cellcolor[HTML]{ACCAEE}\mbox{$-$0.1230} & 0.3724 & \textbf{0.3325} & \cellcolor[HTML]{E4EEFA}\mbox{$-$0.0399} & 0.2623 & \textbf{0.2394} & \cellcolor[HTML]{F0F5FC}\mbox{$-$0.0229} \\
 & CLIP-IQA+$\uparrow$ & 0.3267 & \textbf{0.3517} & \cellcolor[HTML]{ECF3FB}\mbox{+0.0251} & 0.2216 & \textbf{0.3234} & \cellcolor[HTML]{B0CDEF}\mbox{+0.1018} & 0.2219 & \textbf{0.3047} & \cellcolor[HTML]{BFD6F2}\mbox{+0.0827} & 0.2840 & \textbf{0.3404} & \cellcolor[HTML]{D3E3F6}\mbox{+0.0564} & 0.3538 & \textbf{0.4325} & \cellcolor[HTML]{C2D8F3}\mbox{+0.0786} \\
 & MUSIQ$\uparrow$ & \textbf{28.1430} & 26.1142 & \cellcolor[HTML]{FDF4EC}\mbox{$-$2.0288} & 28.1028 & \textbf{34.2600} & \cellcolor[HTML]{CCDFF5}\mbox{+6.1572} & 24.2629 & \textbf{31.9886} & \cellcolor[HTML]{BED6F2}\mbox{+7.7257} & 28.9356 & \textbf{30.5362} & \cellcolor[HTML]{F2F7FC}\mbox{+1.6006} & 44.1707 & \textbf{54.6416} & \cellcolor[HTML]{A8C8ED}\mbox{+10.4709} \\
\midrule
\multirow{5}{*}{\shortstack[l]{\textbf{RawMoment-}\\\textbf{Bench}}} & PSNR$\uparrow$ & 23.5628 & \textbf{26.0642} & \cellcolor[HTML]{95BCEA}\mbox{+2.5014} & 24.8142 & \textbf{27.5100} & \cellcolor[HTML]{8DB7E8}\mbox{+2.6958} & 23.6366 & \textbf{25.3962} & \cellcolor[HTML]{B5D0F0}\mbox{+1.7596} & 24.6596 & \textbf{26.0933} & \cellcolor[HTML]{C2D9F3}\mbox{+1.4337} & 25.4447 & \textbf{26.1280} & \cellcolor[HTML]{E2EDF9}\mbox{+0.6833} \\
 & SSIM$\uparrow$ & 0.7665 & \textbf{0.8228} & \cellcolor[HTML]{B6D1F0}\mbox{+0.0563} & 0.7854 & \textbf{0.8522} & \cellcolor[HTML]{A8C8ED}\mbox{+0.0667} & 0.7563 & \textbf{0.8083} & \cellcolor[HTML]{BBD4F1}\mbox{+0.0520} & 0.7905 & \textbf{0.8269} & \cellcolor[HTML]{D0E1F5}\mbox{+0.0363} & 0.8007 & \textbf{0.8269} & \cellcolor[HTML]{DDE9F8}\mbox{+0.0262} \\
 & LPIPS$\downarrow$ & 0.3143 & \textbf{0.2582} & \cellcolor[HTML]{D9E7F7}\mbox{$-$0.0560} & 0.2634 & \textbf{0.1911} & \cellcolor[HTML]{CEE0F5}\mbox{$-$0.0723} & 0.2825 & \textbf{0.1942} & \cellcolor[HTML]{C3D9F3}\mbox{$-$0.0883} & 0.2584 & \textbf{0.2313} & \cellcolor[HTML]{EDF3FB}\mbox{$-$0.0270} & 0.1850 & \textbf{0.1193} & \cellcolor[HTML]{D3E3F6}\mbox{$-$0.0657} \\
 & CLIP-IQA+$\uparrow$ & 0.3423 & \textbf{0.3619} & \cellcolor[HTML]{F0F5FC}\mbox{+0.0196} & 0.3578 & \textbf{0.3888} & \cellcolor[HTML]{E7F0FA}\mbox{+0.0311} & 0.3453 & \textbf{0.4488} & \cellcolor[HTML]{AECCEF}\mbox{+0.1034} & 0.3469 & \textbf{0.3781} & \cellcolor[HTML]{E7F0FA}\mbox{+0.0312} & 0.4299 & \textbf{0.4788} & \cellcolor[HTML]{D9E7F7}\mbox{+0.0489} \\
 & MUSIQ$\uparrow$ & \textbf{34.0543} & 34.0458 & \cellcolor[HTML]{FFFFFF}\mbox{$-$0.0085} & 38.8111 & \textbf{41.7716} & \cellcolor[HTML]{E6EFFA}\mbox{+2.9606} & 35.9019 & \textbf{43.3717} & \cellcolor[HTML]{C1D8F2}\mbox{+7.4698} & 36.9325 & \textbf{38.5115} & \cellcolor[HTML]{F2F7FC}\mbox{+1.5790} & 51.5117 & \textbf{56.0882} & \cellcolor[HTML]{D9E7F7}\mbox{+4.5765} \\
\midrule
\multirow{2}{*}{\textbf{Magic8-Pro}} & CLIP-IQA+$\uparrow$ & 0.2982 & \textbf{0.3742} & \cellcolor[HTML]{C4DAF3}\mbox{+0.0760} & 0.3054 & \textbf{0.4293} & \cellcolor[HTML]{9FC2EC}\mbox{+0.1238} & 0.3042 & \textbf{0.4507} & \cellcolor[HTML]{8DB7E8}\mbox{+0.1465} & 0.3442 & \textbf{0.4136} & \cellcolor[HTML]{C9DDF4}\mbox{+0.0694} & 0.4012 & \textbf{0.4649} & \cellcolor[HTML]{CDE0F5}\mbox{+0.0637} \\
 & MUSIQ$\uparrow$ & 26.5693 & \textbf{34.2868} & \cellcolor[HTML]{BFD6F2}\mbox{+7.7175} & 30.3885 & \textbf{43.0311} & \cellcolor[HTML]{95BCEA}\mbox{+12.6426} & 30.0586 & \textbf{43.7125} & \cellcolor[HTML]{8DB7E8}\mbox{+13.6539} & 33.5799 & \textbf{40.0396} & \cellcolor[HTML]{C9DDF4}\mbox{+6.4597} & 44.8021 & \textbf{49.0853} & \cellcolor[HTML]{DBE8F8}\mbox{+4.2832} \\
\bottomrule
\end{tabular}
\endgroup
\end{table*}

\paragraph{Experimental setup.}
We train DNF~\citep{jin2023dnf}, LoFormer~\citep{mao2024loformer},
LDM-ISP~\citep{wen2025ldm}, ISPDiffuser~\citep{ren2025ispdiffuser},
and RawMoment separately on the GoPro~\citep{nah2017deep}
and RawMoment training sets.
Evaluation uses the GoPro, RealBlur~\citep{rim2020realblur},
RawMoment-Bench, and Magic8-Pro test sets with the main experiment settings.
We take $512\times512$ center crops of GoPro's blurred RGB images
and apply RawMoment's RAW degradation pipeline and parameter
distributions described in \cref{sec:data,sec:supp_raw_synthesis}.
Sharp RGB crops remain targets; training pairs use three RAW
variants and test pairs one fixed variant.

\paragraph{Mean evaluation metrics.}
\Cref{tab:gopro_dataset_ablation_benchmark} reports image-averaged scores.
Across five methods, training on RawMoment raises PSNR by
1.8148\,dB and SSIM~\citep{wang2004image} by 0.0475 on
RawMoment-Bench, while reducing LPIPS~\citep{zhang2018unreasonable}
by 0.0619. The corresponding changes on RealBlur are
$+0.9342$\,dB, $+0.0396$, and $-0.0814$.

\paragraph{Generalization results.}
For our model, RawMoment training raises CLIP-IQA+~\citep{wang2023exploring}
from 0.3538 to 0.4325 on RealBlur and 0.4012 to 0.4649 on Magic8-Pro;
MUSIQ~\citep{ke2021musiq} increases from 44.1707 to 54.6416 and
44.8021 to 49.0853, respectively. These real-capture gains support
better cross-dataset generalization.

\section{Dataset Diversity Analysis}
\label{sec:supp_diversity}
\setcounter{figure}{0}
\renewcommand{\thefigure}{S\arabic{figure}}
\renewcommand{\theHfigure}{supp.\arabic{figure}}

\paragraph{Features and sampling.}
Frozen DINOv2-small~\citep{oquab2024dinov2} extracts 384-dimensional
CLS features from clean training RGB images.
We uniformly sample 1,024 images without replacement per dataset
for equal-count comparisons; GoPro is the smallest with 2,103 images.

\begin{figure*}[p]
\centering
\includegraphics[width=.88\textwidth]{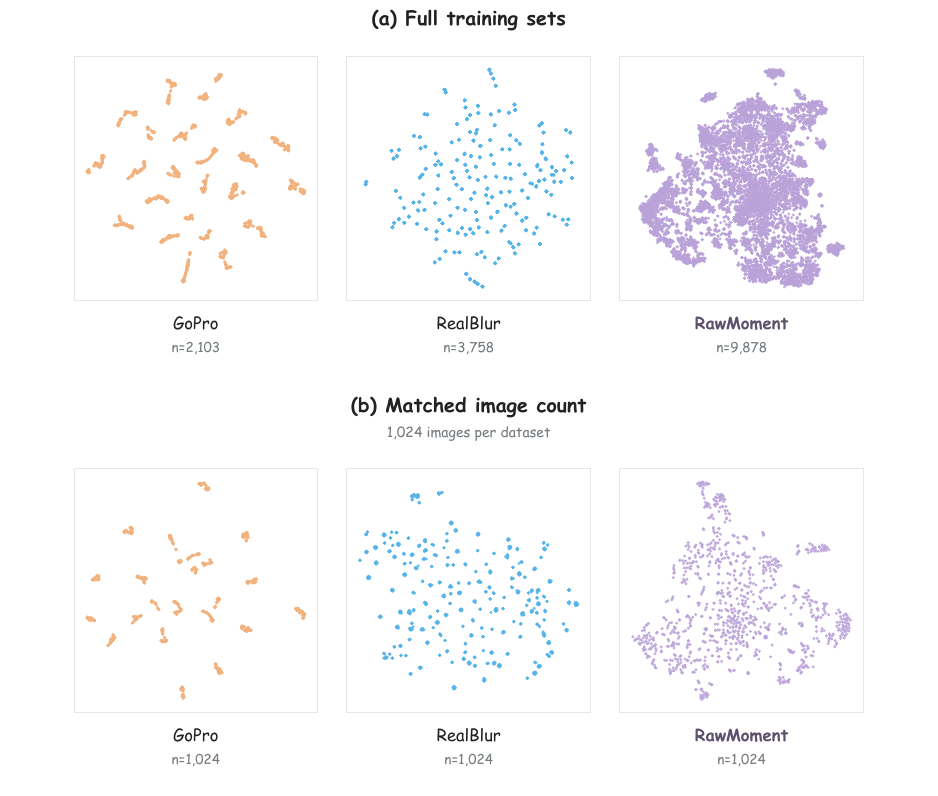}
\caption{t-SNE projections of normalized DINOv2-small features.
(a) Full clean training sets; (b) random 1,024-image subsets.
Each t-SNE panel is fitted independently;
coordinates and occupied areas are not directly comparable
across panels.}
\label{fig:supp_tsne}
\par\vspace{8pt}
\includegraphics[width=\textwidth]{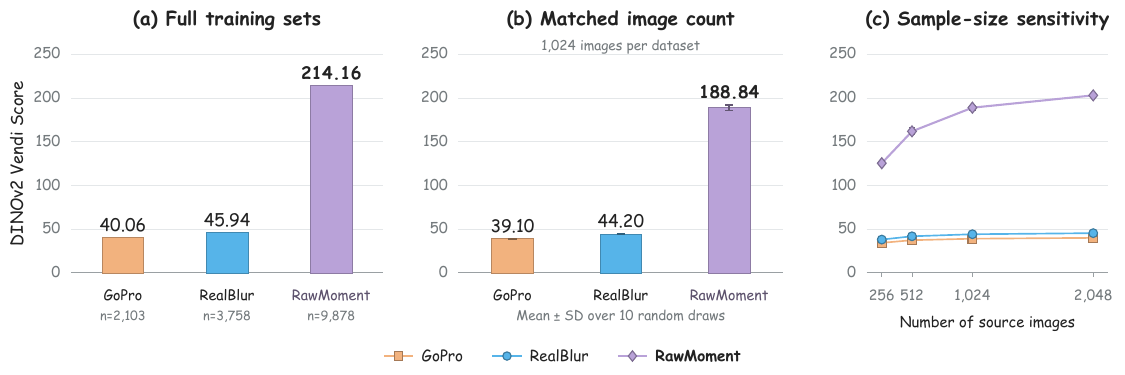}
\caption{Visual feature diversity measured by the Vendi score.
(a) Full clean training sets; (b) matched subsets of 1,024 images;
(c) sample-size sensitivity.
For (b) and (c), each draw samples without replacement from the
full training set; error bars show the standard deviation over
ten draws. RAW degradation variants are not counted as
additional source frames.}
\label{fig:supp_vendi}
\end{figure*}

\paragraph{t-SNE visualization.}
We fit t-SNE~\citep{vandermaaten2008tsne} independently for each
panel in \cref{fig:supp_tsne}, using perplexity 30, 1,500 iterations,
Euclidean distance, an automatic learning rate, and seed 42.
The 1,024-image subsets are drawn with sampling seed 0.

\paragraph{Visual feature diversity.}
For $n$ unit features $\mathbf{f}_i$, the uncentered cosine kernel
is $K_{ij}=\mathbf{f}_i^{\top}\mathbf{f}_j$.
The Vendi score~\citep{friedman2023vendi} is
$\mathrm{VS}=\exp(-\sum_k\lambda_k\log\lambda_k)$, where
$\lambda_k$ are the eigenvalues of $K/n$ and $0\log0=0$.
\Cref{fig:supp_vendi} shows mean scores over ten 1,024-image draws:
188.84 for RawMoment, 44.20 for RealBlur, and 39.10 for GoPro.
The ranking holds across the tested subset sizes.
The greater feature diversity and the generalization gains in
\cref{sec:supp_training_data} support RawMoment as diverse and
effective training data.

\section{Additional Qualitative Results}
\label{sec:supp_qualitative}
\begin{figure*}[p]
\centering
\includegraphics[width=\textwidth,height=.82\textheight,keepaspectratio]{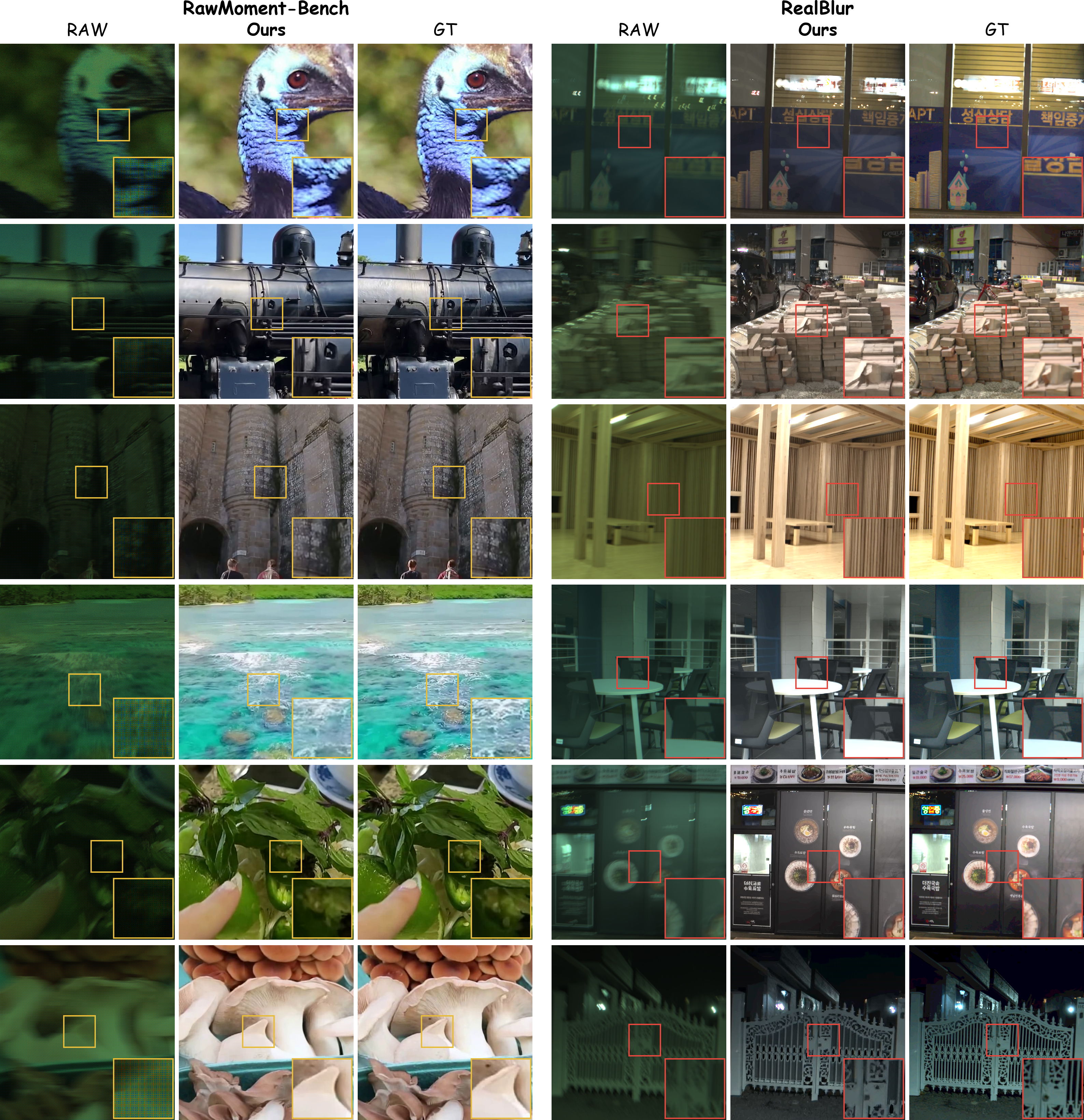}
\caption{Additional qualitative results on RawMoment-Bench (left)
and RealBlur (right). Each group shows the RAW input, our restored
RGB image, and the paired ground truth for six scenes. Yellow
(RawMoment-Bench) and red (RealBlur) boxes mark the corresponding
regions enlarged in the lower-right insets.}
\label{fig:supp_qualitative}
\end{figure*}

\begin{figure*}[p]
\centering
\includegraphics[width=\textwidth,height=.82\textheight,keepaspectratio]{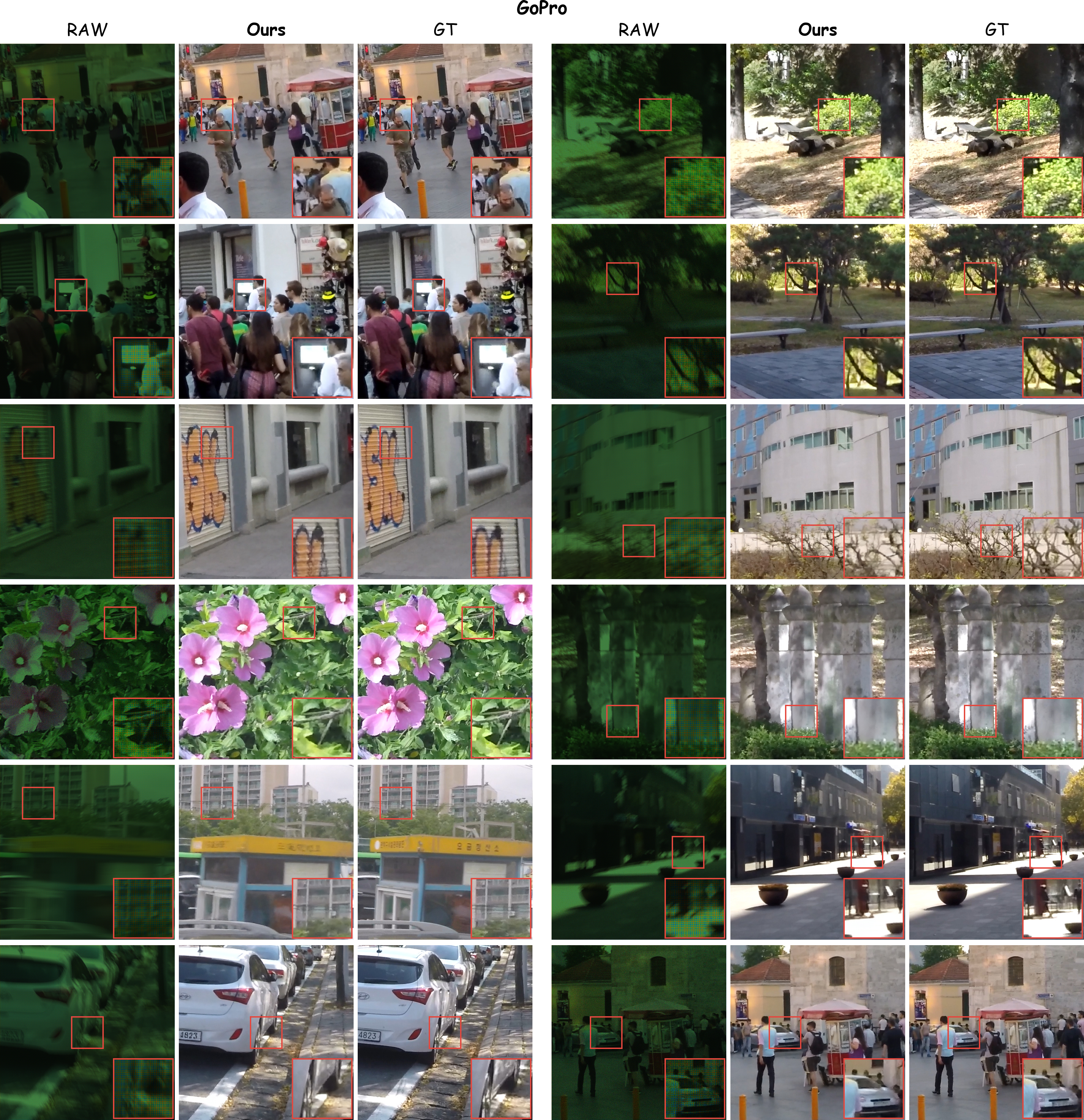}
\caption{Additional GoPro results in two groups of six scenes,
each comparing the RAW input, our restored RGB image, and the
paired ground truth. Red boxes mark
the matched regions enlarged in the lower-right insets.}
\label{fig:supp_gopro}
\end{figure*}

Figures~\ref{fig:supp_qualitative} and~\ref{fig:supp_gopro}
present additional qualitative results on RawMoment-Bench,
RealBlur, and GoPro. Each group compares RAW inputs, restored
RGB images, and paired references.
RealBlur visualizations use the same scalar brightness
alignment~\citep{rim2020realblur} as the main evaluation.

\clearpage

\clearpage

\end{document}